\documentclass[10pt,journal,compsoc]{IEEEtran}

\usepackage{cite}
\usepackage{multirow}
\usepackage{amssymb}
\usepackage{xspace}
\usepackage{epsf}
\usepackage{setspace}
\usepackage{caption}
\usepackage{comment}
\usepackage{color}
\usepackage{amsmath}

\def\method{CORE\xspace}

\usepackage[ruled,vlined,linesnumbered,ruled]{algorithm2e}
\usepackage{balance}
\usepackage{url,epsfig, array}
\usepackage{amsmath}
\usepackage{epstopdf}
\usepackage{cases}
\usepackage{bm}
\usepackage{color}
\usepackage[table]{xcolor}
\usepackage{tikz}
\usepackage{threeparttable}

\usepackage{diagbox}

\usepackage{microtype}
\usepackage{graphicx}
\usepackage{booktabs} 
\usepackage{subcaption}
\usepackage{times} 
\usepackage{balance}
\usepackage{enumitem}
\usepackage{multirow}
\usepackage{bbm}
\usepackage{tablefootnote}
\usepackage{threeparttable}
\usepackage{xspace}
\usepackage{amsmath}
\usepackage{amsfonts}
\usepackage{amssymb}
\usepackage{bm}
\usepackage{makecell}
\usepackage{setspace}
\usepackage{amsmath}
\usepackage{amssymb}
\usepackage{mathtools}
\usepackage{amsthm}
\usepackage{nomencl}
\usepackage{multirow}
\usepackage{adjustbox}
\usepackage{multirow}

\newcommand*{\circled}[1]{\lower.8ex\hbox{\tikz\draw (0pt, 0pt) circle (.47em) node {\makebox[0.4em][c]{\small #1}};}}

\usepackage{pifont} 

\def\ie{\textit{i.e.}\xspace}

\def\eg{\textit{e.g.}\xspace}

\begin{document}
	\title{Less is More: Lightweight Prompt Compression for Question Answering Applications on Edge Devices}
	\author{
		\IEEEauthorblockN{
			Zihuai Xu,~
            Ruofei Hou,~
            {*}Yang Xu,~\IEEEmembership{Member,~IEEE,}~
            Hongli Xu,~\IEEEmembership{Member,~IEEE,}~
            Yunming Liao,~
            Ying Zhu\\
		}
        \IEEEcompsocitemizethanks{
            \IEEEcompsocthanksitem This article is supported in part by the National Science Foundation of China (NSFC) under Grants XXXXXXXX, XXXXXXXX, and XXXXXXXX; in part by the Fundamental Research Funds for the Central Universities (Grant No. WKXXXXXXXXXX).
			\IEEEcompsocthanksitem Z. Xu, R. Hou, Y. Xu, H. Xu, Y. Liao, and Y. Zhu are with the School of Computer Science and Technology, University of Science and Technology of China, Hefei, Anhui, China, 230027, and also with Suzhou Institute for Advanced Research, University of Science and Technology of China, Suzhou, Jiangsu, China, 215123. \protect E-mails: zihuaixu@mail.ustc.edu.cn; xuyangcs@ustc.edu.cn; xuhongli@ustc.edu.cn; ymliao98@mail.ustc.edu.cn; zhiweiyao, \{zhiweiyao, isaaczhu\}@mail.ustc.edu.cn.
		}
	}

	\markboth{IEEE Transactions on XXX, Vol., No., May. 2025}%
	{Shell \MakeLowercase{\textit{et al.}}: Bare Advanced Demo of IEEEtran.cls for Journals}
	
	\IEEEtitleabstractindextext{%
		\begin{abstract}
			In agent-driven question answering (QA) applications, retrieval-augmented generation (RAG) is commonly introduced to enhance the response accuracy of large language models (LLMs) by providing additional context. Due to the inherent noise in retrieval results and the coarse granularity of document-level retrieval, the retrieved context often contains substantial redundant information. In this setting, the agent prompt, consisting of the user query and the associated retrieved context, leads to unnecessary computational overhead during LLM inference.
Existing prompt compression methods typically rely on auxiliary small language models (SLMs) to estimate context importance. However, such approaches introduce significant memory and computational overhead, which limits their deployment on resource-constrained edge devices.
In this paper, we propose \textit{COntext Refinement via Entity-aware filtering} (\method), a two-stage sentence-level prompt compression method that eliminates the need for SLMs. In the first stage, \method constructs an answer set via named entity recognition (NER) and a clue set via semantic matching. In the second stage, \method refines the clue set using an orthogonal residual retrieval strategy and designs a spatial proximity-based metric to filter the answer set. The two sets are then combined to form the final compressed context.
We implement \method on an NVIDIA Jetson AGX Orin edge device and a Huawei Nova smartphone. Experimental results demonstrate that within a 2000-token budget, \method improves accuracy by at least 30.19\% compared to state-of-the-art baselines, while reducing memory usage by at least 50.47\% and achieving at least 1.94$\times$ speedup on the edge device. Moreover, compared to the state-of-the-art LLMLingua2 method, \method achieves a substantial energy reduction of 95.74\% on the smartphone, highlighting its practicality and generalizability for mobile deployments.

		\end{abstract}
	}
	
	\maketitle
	\IEEEdisplaynontitleabstractindextext
	\IEEEpeerreviewmaketitle
	
	\section{Introduction}
	Large language model (LLM) powered agents are evolving at an unprecedented pace \cite{barua2024exploring, wang2025large}.
Their deployment has expanded from cloud-centric infrastructures to edge devices, such as smartphones and autonomous vehicles \cite{sun2025automated, bhat2025llm}.
Among agents' diverse capabilities, question answering (QA) stands out as the primary gateway through which users access a vast amount of information.
The interaction paradigm of QA is shifting from brief exchanges to lengthy tasks involving extensive contexts, such as retrieving specific details from historical archives \cite{zuo2025rise, virodhula2025ai}.
However, the inherent limitations of LLMs, including outdated training data \cite{zhang2023large, cheng2024dated} and a lack of domain-specific expertise \cite{ling2023domain, song2025injecting}, pose a persistent risk of hallucinations \cite{lan2024survey, tonmoy2024comprehensive, huang2025survey}, \ie, the generation of plausible yet incorrect responses.
To mitigate the hallucinations of LLMs, retrieval-augmented generation (RAG) has emerged as an indispensable technique \cite{yu2024evaluation, zhao2024retrieval}.
By retrieving relevant documents from a real-time knowledge base, RAG enables agents to ground their responses in external evidence rather than relying solely on parametric knowledge, thereby ensuring the delivery of reliable answers for applications \cite{zhang2024enhancing, mohammad2024advancing}.

Despite its effectiveness in grounding answers, RAG in practical QA applications suffers from the long-context problem \cite{jinlong, laban2024summary}.
Empirical evidence from \cite{jiang2023llmlingua} demonstrates that in standard RAG configurations, the proportion of redundant context can exceed 95\%.
Such noise dilutes key information, hindering the extraction of correct answers and sometimes degrading response accuracy paradoxically \cite{baker2024lost, zhang2025lost}.
Moreover, due to the quadratic complexity of self-attention with respect to input length, the increased prompt length, \ie, the query alongside the context, can significantly intensify LLM inference overhead, delaying responses in real-time QA systems. 
Furthermore, excessively long contexts may even exceed the LLM's context window limit \cite{lenglong, li2025long}. 
For example, retrieved contexts frequently exceed 10k tokens, presenting a mismatch with the constrained context windows of many edge-side models, such as the \texttt{LLaMA2-7B-chat-4k} model \cite{touvron2023llama}.


Prompt compression has become a key strategy for mitigating the long-context bottleneck by constraining the input to a fixed token budget while preserving essential information. 
A subset of compression approaches fine-tunes auxiliary small language models (SLMs) as filters to identify and remove redundant content.
However, applying these methods in edge environments presents significant challenges.
For instance, \cite{liskavets2025prompt} involves fine-tuning the \texttt{Mistral-7B}~\cite{jiang2023mistral}, a process that demands computational capabilities on par with, or exceeding, those of an RTX 3090-tier GPU.
Although \cite{pan2024llmlingua} lowers the computational barrier by employing a lighter 1B backbone based on \texttt{XLM-RoBERTa-Large}~\cite{conneau2020xlmr}, it remains hindered by a heavy reliance on extensive training data (usually $>$50k samples). 
Such training-based methods require extensive resources, making them unsuitable for edge scenarios.

\begin{figure}[t]
\centering
\includegraphics[height=5cm]{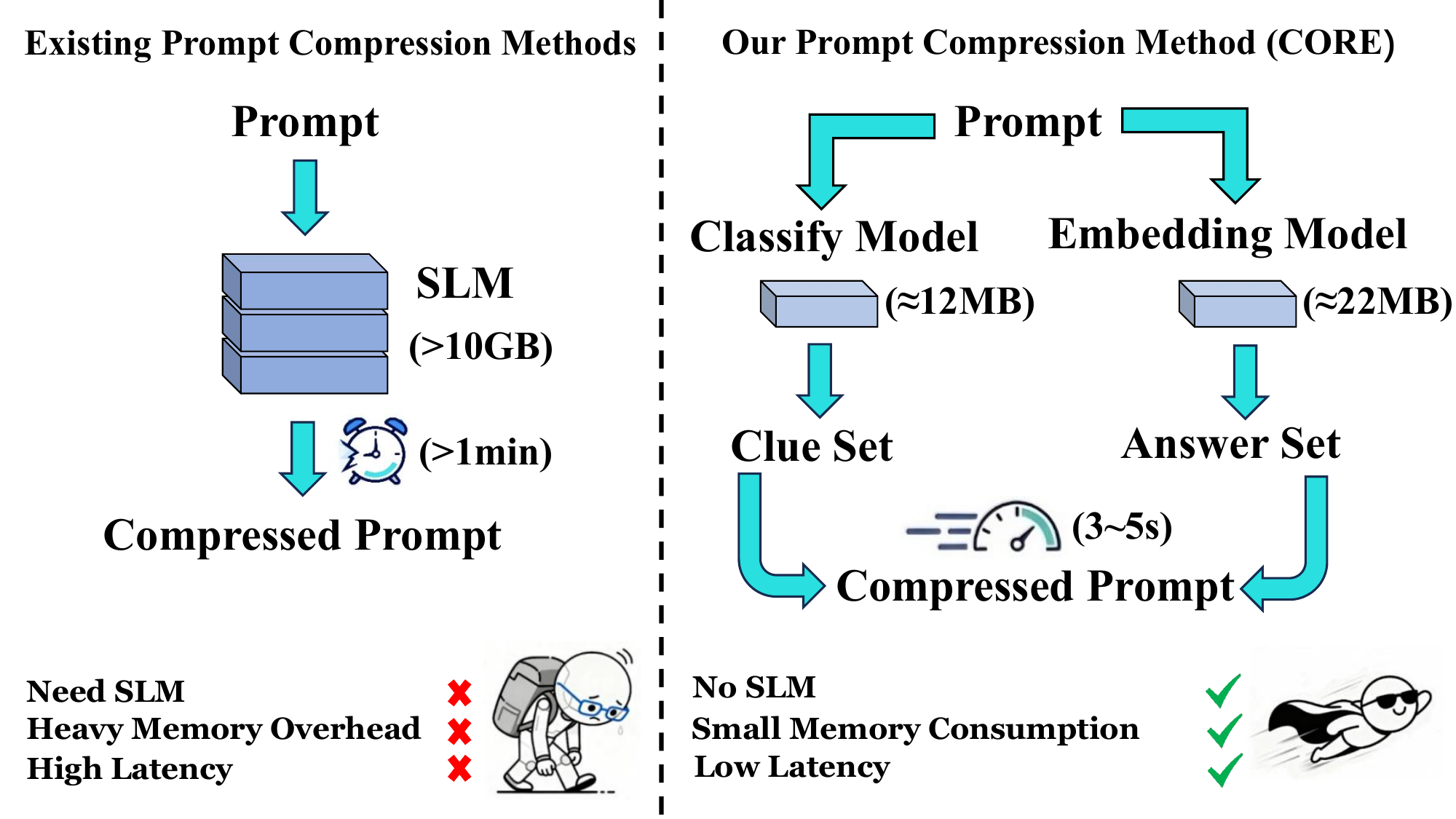}
\caption{Three distinct query cases in real QA tasks alongside their corresponding interested entities.}
\label{fig:Simpleoverview}
\end{figure}


In contrast, training-free approaches, which directly utilize pretrained SLMs to guide compression, have become mainstream. 
To avoid fine-tuning, these methods rely on relatively capable SLMs (typically with around 7B parameters) to compute different metrics for assessing the importance of tokens in the context. 
Early works like \cite{li2023unlocking} employ self-information \cite{shannon1948mathematical}, \ie, the negative log probability of a token, as the importance metric and utilize SLMs, \eg, the \texttt{Curie-6.7B} model \cite{li2023unlocking}, to compute token importance. 
Subsequently, perplexity-based methods gain prominence. 
Their principle is that tokens with lower perplexity on SLMs contribute less value, pruning these tokens will not significantly affect the response quality.
For example, both \cite{jiang2023llmlingua} and \cite{jiang2024longllmlingua} are based on the \texttt{LLaMA2-7B-hf} model~\cite{touvron2023llama} to compute token perplexity.
More recently, attention-driven methods have emerged as a promising direction, where the attention module in SLMs is considered the key to determining a token‘s importance. 
For instance, \cite{fei2025efficient} leverages specific attention heads in the \texttt{LLaMA-3.1-8B} model \cite{kassianik2025llama} to compute token attention scores and removes tokens with lower attention scores within a token budget.

However, current training-free approaches exhibit notable limitations when deployed on edge devices. 
Concretely, 1) \textbf{\textit{Heavy Memory Overhead}}. 
Auxiliary SLMs typically consume substantial memory resources. 
For example, a 7B SLM under common half-float storage occupies 14 GB for model weights alone, and can demand up to 24 GB of memory during runtime. 
This requirement far exceeds the typical 8–16 GB of GPU VRAM available on consumer edge devices \cite{nvidia2022jetsonorinnx, you2024edgellm, xu2025edgellm}. 
2) \textbf{\textit{High Computation Latency}}. 
Both perplexity and self-attention computations require prefill of the prompt, a compute-intensive process that introduces notable latency on edge devices with constrained computing power \cite{sheng2023flexgen}. 
For instance, prefilling 10k tokens with a 3B SLM on a smartphone results in delays exceeding one minute \cite{xu2024mllmnpu}. Such prolonged latency is unacceptable for time-sensitive scenarios like real-time QA.

%

Driven by the insight that QA answers are typically anchored to specific entities \cite{grishman1996message}, \ie, key information carriers such as dates and locations, we propose \method, a lightweight two-stage prompt compression framework. 
In the first stage, \method performs candidate filtering. 
It leverages an entity classifier specifically designed for named entity recognition (NER) together with an embedding model as a coarse-grained filter to construct an answer set, \ie, sentences that potentially contain the answer, and a clue set, \ie, sentences that are highly relevant to the query. 
Owing to their functionally specialized design, both models are very lightweight, with their total weights occupying $\approx$ 40 MB of memory, which is merely 0.14\% of the weight memory of a 7B SLM. 
Building on the filtered results, \method then enters the evidence refining stage. 
It exploits the characteristics of the sentence sets constructed in the first stage to distill a concise context efficiently. 
Specifically, to refine the clues, \method employs an orthogonal vector decomposition strategy to compress the clue set by minimizing semantic redundancy. 
Meanwhile, \method introduces a computationally efficient metric to filter the answer set based on entity density. 
By eliminating the reliance on SLMs, \method ensures negligible memory overhead and low latency on edge devices.


However, refining these two interdependent subsets within a rigid token budget presents a nontrivial challenge.
The difficulty stems from the intrinsic connection between the two sets: the utility of a clue is largely contingent upon its ability to bridge specific answers to the query, while the validity of a candidate answer relies heavily on the spatial support provided by high-confidence clues.
Mere independent selection risks retaining disjointed information, where excessive clues may introduce noise, or insufficient evidence could break reasoning chains, ultimately resulting in hallucinations.

In summary, our main contributions are as follows:
\begin{itemize}
    \item We propose \method, a lightweight two-stage framework for compressing prompts in QA applications, and we design a candidate filtering mechanism that synergizes NER with semantic matching to extract an answer set and a clue set. 
    By eliminating the reliance on SLMs, we achieve reductions in both memory overhead and computation latency.
    \item We introduce an answer-aware method to discard irrelevant clues, and devise a metric based on spatial proximity to clues to filter out distracting answers, thereby intertwining the refinement of both sets.
	\item We conduct extensive experiments on an NVIDIA Jetson edge device. Results demonstrate that \method improves accuracy by at least 30.19\% compared to the baseline methods, while reducing memory usage and compression latency by at least 50.47\% and 48.52\%, respectively. Meanwhile, \method achieves an energy reduction of 95.74\% compared to the state-of-the-art LLMLingua2 method on a Huawei Nova smartphone.
\end{itemize}

The remainder of this paper is organized as follows: Section \ref{sec:prelim} introduces the background of prompt compression. Sections \ref{sec:motivations} and \ref{sec:algorithm} elaborate on our motivations and detailed \method, respectively. Section \ref{sec:evaluation} presents extensive experimental results. Finally, Section \ref{sec:conclusion} concludes the paper, and Section \ref{sec:relatedwork} reviews related work.
	
	\section{Preliminaries}\label{sec:prelim}

\subsection{Augmented Prompt}
In RAG-enhanced QA applications, the augmented prompt serves as the fundamental input of LLMs.
Formally, an augmented prompt $x$ consists of a specific user query $x_{\text{q}}$ and a supporting context $x_{\text{s}}$.
The query $x_{\text{q}}$ encapsulates the user's inquiry, typically seeking a precise fact.
The context $x_{\text{s}}$ provides the external knowledge required to extract the correct answer, which is typically obtained by retrieving the top-$k$ relevant document chunks about $x_{\text{q}}$.
We denote the retrieved context as a sequence of $m$ sentences, $x_{\text{s}} = \{s_1, s_2, \dots, s_m\}$, where each sentence $s_i$ consists of $n_i$ tokens.
While enriching the input with necessary evidence, the direct concatenation of retrieved documents often introduces significant redundancy and irrelevant noise \cite{yoran2024making}.
In QA tasks, this noise may exceed the context window capacity of LLMs and attenuate attention allocation to key evidence~\cite{liu2024lost}.

\subsection{Prompt Compression}
To address the aforementioned challenges, prompt compression seeks to extract essential information from the augmented prompt while satisfying computational resource constraints without compromising answer accuracy.
Given an input prompt $x = (x_{\text{s}}, x_{\text{q}})$, the goal is to transform the original context $x_{\text{s}}$ into a compressed version $\hat{x}_{\text{doc}}$ while retaining critical evidence that supports the ground-truth answer.
We formulate the objective of this compression task as maximizing the utility of the LLM's output:
\begin{equation}\label{obj1}
	\max_{\hat{x}} \Phi \Big[ \text{LLM}(\hat{y} \mid \hat{x}) \Big]
\end{equation}
where $\text{LLM}(\hat{y} \mid \hat{x})$ represents the generation probability of the answer $\hat{y}$ given the compressed prompt $\hat{x}$, and $\Phi(\cdot)$ serves as the utility evaluator, specifically the prediction accuracy, \eg, accuracy, F1-Score, ROUGE and Perplexity.

In general, prompt compression can operate at two primary granularities, \ie, token and sentence.
Compared to token-level compression that discards tokens individually, the sentence-level approach better preserves the semantic integrity and grammatical correctness of the evidence sentences, as it avoids disrupting the underlying semantic structure \cite{liskavets2025prompt}.
Therefore, we focus on sentence-level context compression in this work.
Consequently, given a strict token budget constraint $\mathcal{B}$, the optimization problem in Eq.~\eqref{obj1} implies selecting a subset of sentences that maximizes answer accuracy while satisfying the constraint:
\begin{align}\label{obj2}
	~~~~~~~~~&\max_{\hat{x}_{\text{s}}} \Phi \Big[ \text{LLM}(\hat{y} \mid (\hat{x}_{\text{doc}}, ~ x_{\text{q}})) \Big], \\
	\nonumber &\text{s.t.}\, \sum_{s_i \in \hat{x}_{\text{s}}} n_i + len(x_{\text{q}})\leq \mathcal{B} 
\end{align}
where $len(x_\text{q})$ represents the number of tokens contained in $x_\text{q}$.
The ultimate goal is to generate a compressed prompt $\hat{x}$ that is sufficiently concise to minimize inference latency and memory usage, yet sufficiently informative to remain faithful to the original retrieved evidence for correct reasoning.

    \section{Insights and motivation}\label{sec:motivations}
This section begins by analyzing the defining characteristics of QA tasks, revealing that user queries demonstrate specific intents targeting distinct entity types (Section~\ref{sec:insights}). 
Building on this insight, we introduce a fast prompt compression approach that constructs an answer sentence set based on entity analysis (Section~\ref{sec:answer-set}) and a clue sentence set via semantic matching (Section~\ref{sec:clue-set}).
We demonstrate its effectiveness via preliminary experiments (Section~\ref{sec:motivation}).

\subsection{Insights}\label{sec:insights}
\begin{figure}[t]
\centering
\includegraphics[height=5.5cm]{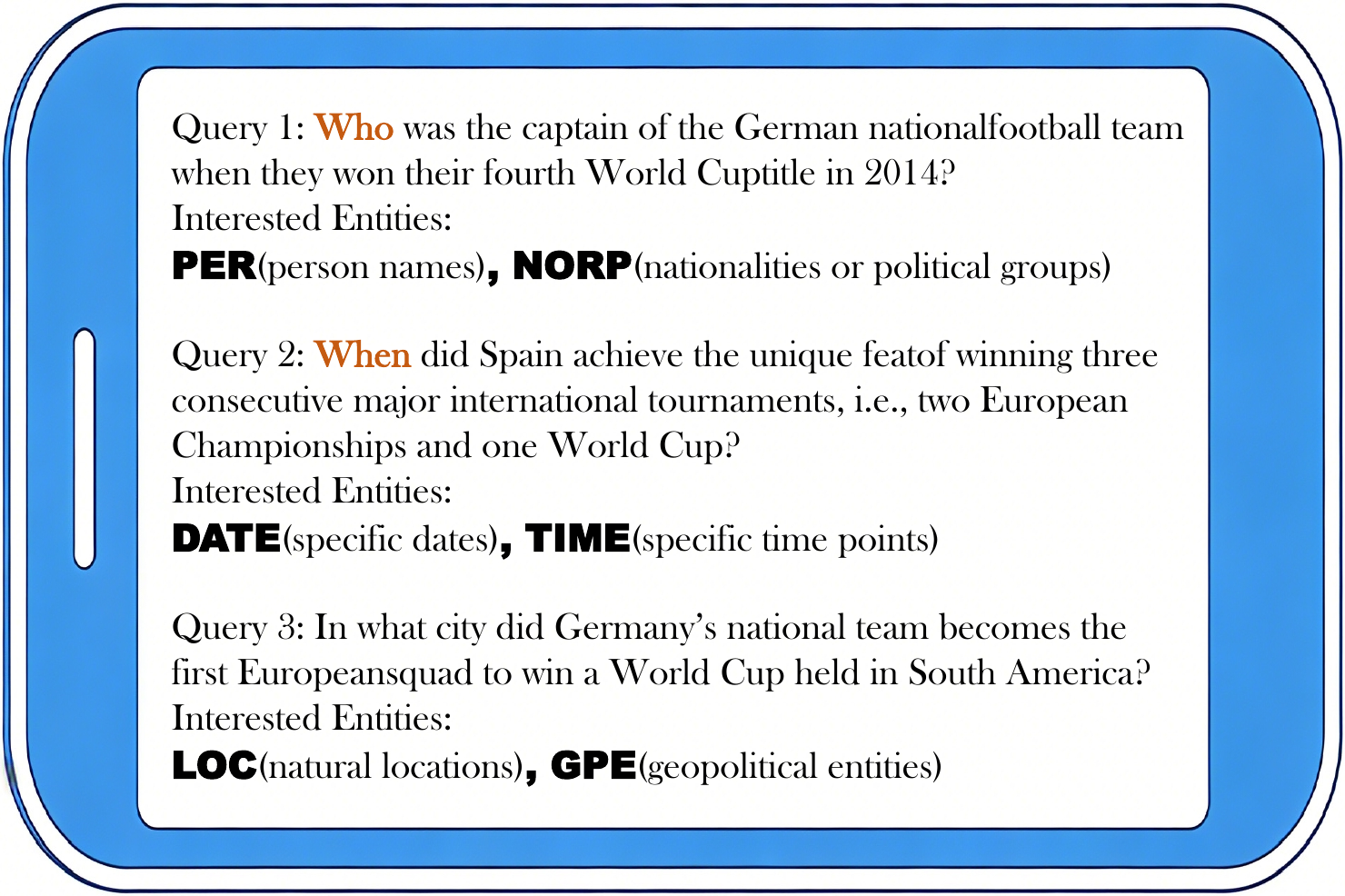}
\caption{Three distinct query cases in real QA tasks alongside their corresponding interested entities.}
\label{fig:EXAns}
\vspace{-0.3cm}
\end{figure}

In QA tasks, user queries consistently demonstrate specific intents targeting distinct entity types.
As depicted in Figure~\ref{fig:EXAns}, different queries have unique preferences for particular entity categories.
Since the truth answer is invariably rooted in these specific entities, a candidate answer set, \ie, sentences that contain at least one of the interested entities, can be effectively derived from the context.

Recognizing the set of interested entities $\mathbb{\hat{E}}$ for the query $x_{\text{q}}$ constitutes a prerequisite for accurately constructing the answer set.
A naive recognizing method is mapping interrogative pronouns to specific entities~\cite{sasazawa2019neural}.
For example, as shown in Figure~\ref{fig:EXAns}, the interrogative word ``\textit{Who}'' in Query 1 explicitly signals an intent for person-related entities, \eg, \(\mathbf{PER}\) and \(\mathbf{NORP}\).
Similarly, ``\textit{When}'' in Query 2 points to temporal entities such as \(\mathbf{DATE}\) and \(\mathbf{TIME}\).
However, this approach proves brittle when handling indirect or complex queries.
As exemplified by Query 3, interrogative words like ``\textit{What}'' lack specialization and cannot be associated with any specific entity type.
Consequently, a more robust method is required to accurately infer the entity intent underlying diverse query formulations.


\subsection{Answer Set}\label{sec:answer-set}
We introduce an intent analysis mechanism grounded in semantic matching.
Specifically, we define a set of canonical template queries, where each template is meticulously designed and corresponds to a specific category of entities derived from linguistic knowledge.
Representative examples include ``\textit{When did this event occur?}'' for temporal contexts, ``\textit{Who is the person of interest?}'' for identity-focused queries, and ``\textit{Where is the place?}'' for location-based inquiries.
The intent of \(x_{\text{q}}\) is subsequently determined by measuring its semantic similarity to these templates within the high-dimensional embedding space.
The entity categories associated with the template yielding the highest similarity score constitute the interested entity set \(\mathbb{\hat{E}}\).
For instance, despite lacking explicit directional keywords, Query 3 in Figure~\ref{fig:EXAns} exhibits the highest semantic affinity with the location query template, thereby correctly associating it with \(\mathbf{LOC}\) and \(\mathbf{GPE}\) entities.

Formally, let $\mathbb{Q} = \{q_1, q_2, \dots\}$ be the set of template queries, where each template $q_i$ maps to a predefined set of entities $\mathbb{E}_i=\{e^i_1, e^i_2, \dots\}$.
The semantic similarity between the user query $x_{\text{q}}$ and a template $q_i$ is computed as:
\begin{equation}\label{sim}
	\sigma(x_{\text{q}}, q_i) = M_{\text{emb}}(x_{\text{q}}) \cdot M_{\text{emb}}(q_i)
\end{equation}
where $M_{\text{emb}}(\cdot)$ denotes the semantic vector representation generated by the embedding model.
Therefore, the interested entity set $\mathbb{\hat{E}}$ is identified as the set corresponding to the template with the maximum similarity:
\begin{equation}\label{Ien}
	\mathbb{\hat{E}} = \mathbb{E}_{j}, \quad \text{where } j = \text{argmax}_{i} \big[ \sigma(x_{\text{q}}, q_j) \big]
\end{equation}

\begin{figure}[t]
\centering
\includegraphics[height=5.5cm]{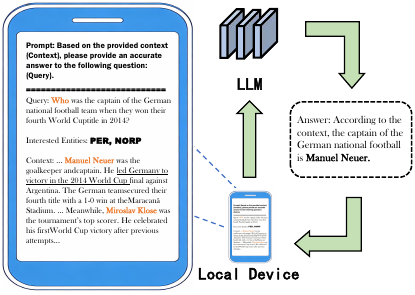}
\caption{An example of a query with RAG contextual clues and the correct answer. The clue sentence is highlighted with underscores.}
\label{fig:EXAns2}
\vspace{-0.4cm}
\end{figure}

Leveraging the lightweight entity classifier $M_{\text{cls}}$, \ie, a model that extracts both entities and their corresponding types from input sentences, we obtain $M_{\text{cls}}(s_j)$, which represents the set of entities contained within sentence $s_j$.
Based on this, the candidate answer set $\mathbb{C}_{\text{ans}}$ is constructed by filtering for sentences in the context $x_{\text{s}}$ that contain at least one entity belonging to $\mathbb{\hat{E}}$:
\begin{equation}\label{ansset}
	\mathbb{C}_{\text{ans}} = \{ s_j \in x_{\text{s}} \mid M_{\text{cls}}(s_j) \cap \mathbb{\hat{E}} \neq \emptyset \}
\end{equation}
This semantic matching strategy enables robust identification of user intent.
By filtering context based on semantic interest, this process significantly minimizes information redundancy.
Nevertheless, merely retaining $\mathbb{C}_{\text{ans}}$ is insufficient to formulate a complete and coherent response, as the model also requires supporting context to link these potential answers to the query.
Therefore, additional contextual clues are required to facilitate accurate reasoning.

\begin{figure*}[t]
	\hspace*{-0.3cm}
    \begin{subfigure}{0.31\textwidth}
		\centering
		\includegraphics[height=4.cm]{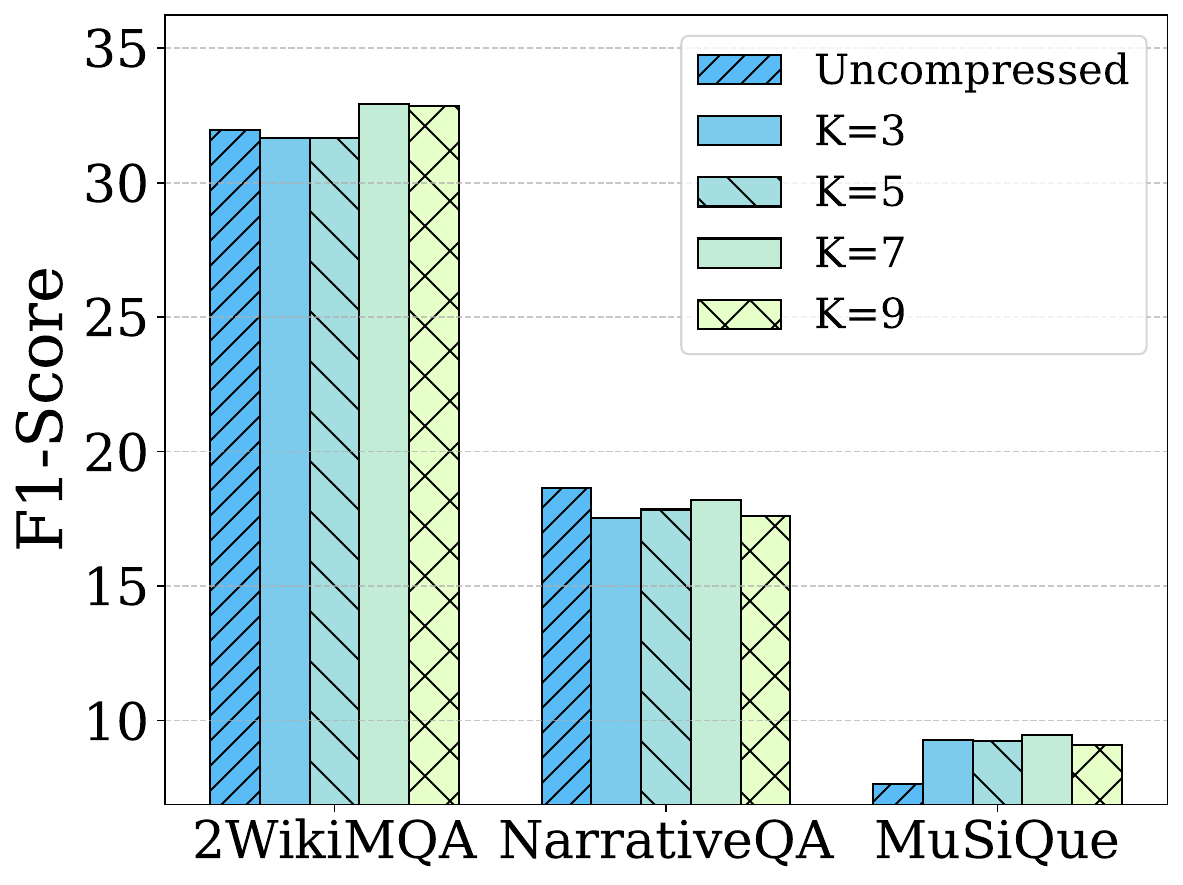}
		\caption{LLaMA2-7B-chat-4k}
		\centering
		\label{fig:llama2_7b_chat_pre}
	\end{subfigure}
	\hspace{0.03\textwidth}
	\begin{subfigure}{0.31\textwidth}
		\centering
		\includegraphics[height=4.cm]{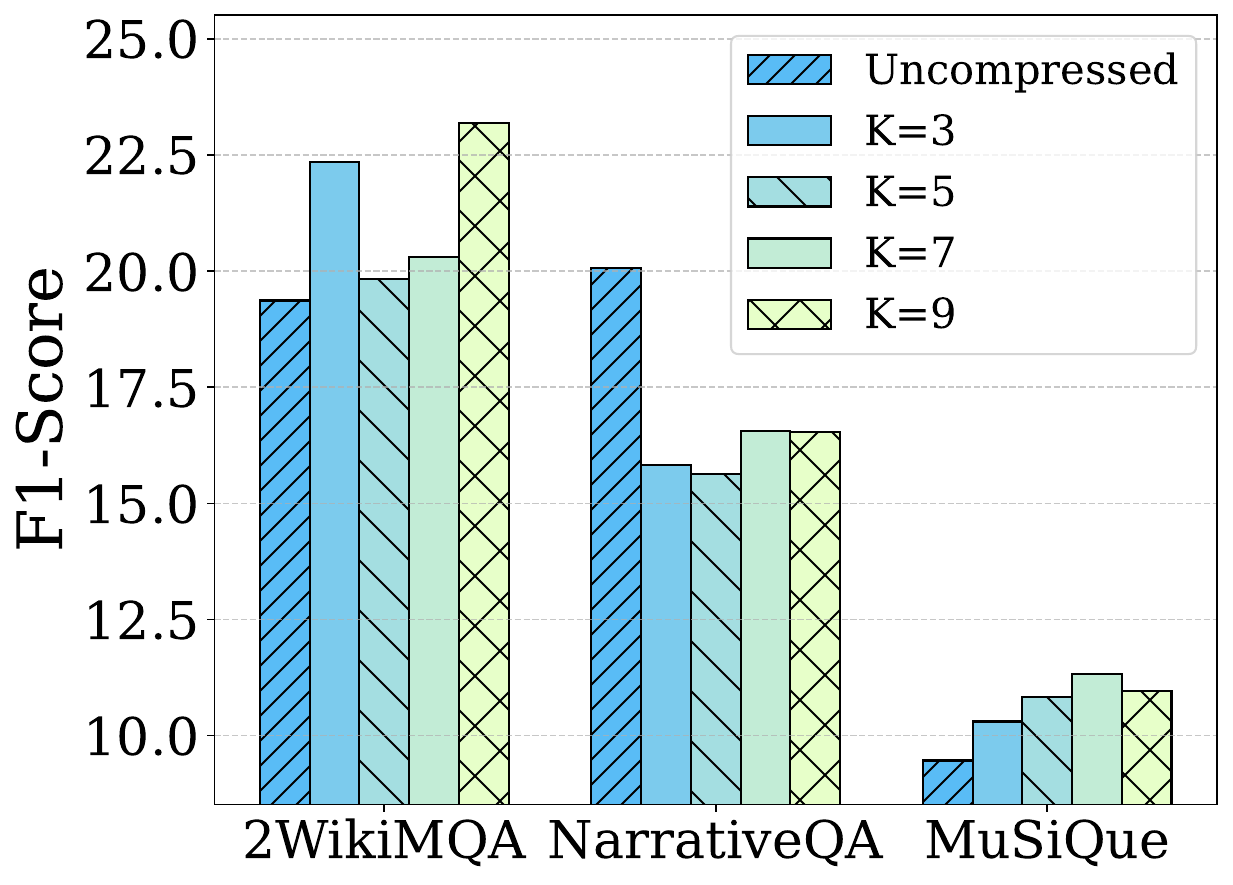}
		\caption{Longchat1.5-7B-32k}
		\centering
		\label{fig:longchat_v1_5_7b_pre}
	\end{subfigure}
	\hspace{0.03\textwidth}
	\begin{subfigure}{0.31\textwidth}
		\centering
		\includegraphics[height=4.cm]{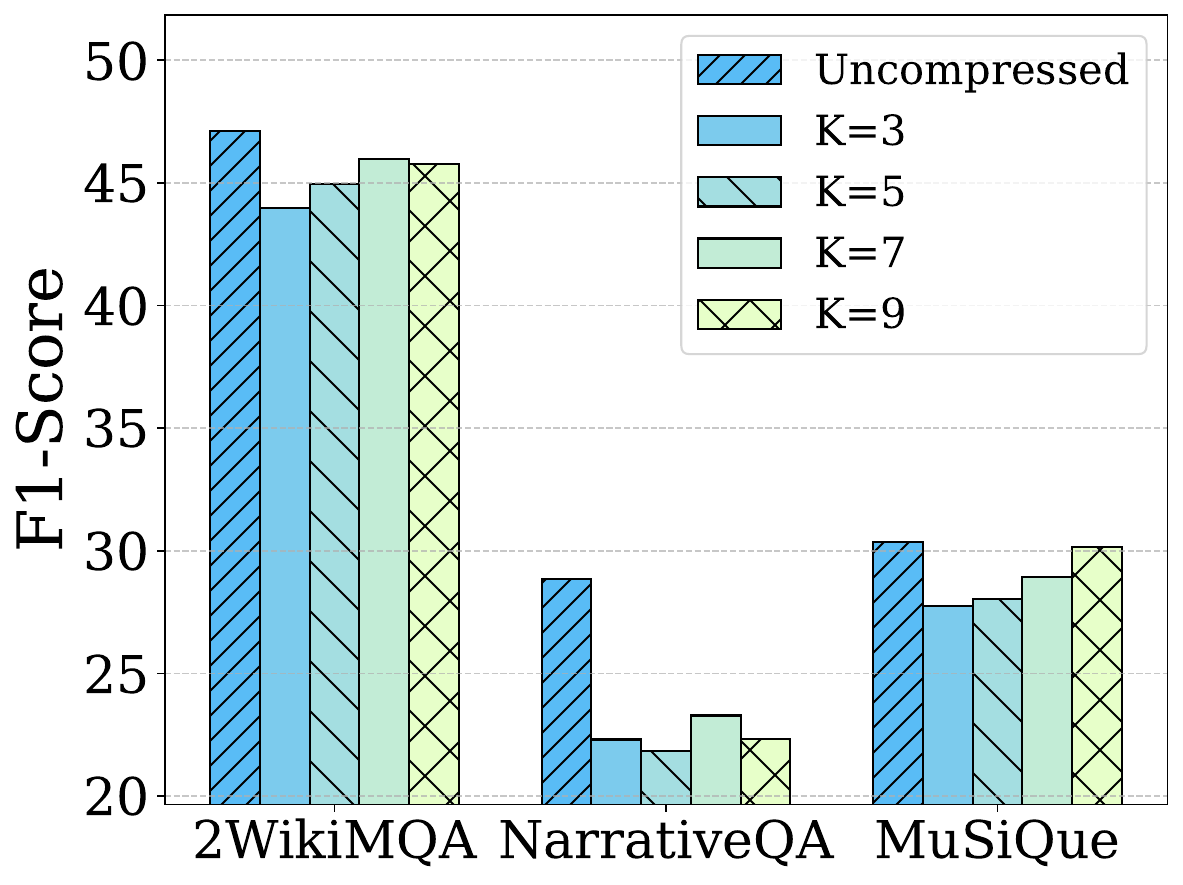}
		\caption{Qwen2.5-7B-Instruct}
		\label{fig:qwen2_7b_pre}
	\end{subfigure}
	\caption{Performance variation with clue set size $K$ across different models on three LongBench datasets.}
	\label{fig:PreEX}
    \vspace{-0.3cm}
\end{figure*}

\subsection{Clue Set}\label{sec:clue-set}
The candidate answer set $\mathbb{C}_{\text{ans}}$ frequently encompasses multiple potential answers due to the extensive length of retrieved documents $x_{\text{s}}$ and the prevalence of multiple entities within the text.
This often leads to ambiguity regarding the ground-truth answer.
To effectively pinpoint the correct entity among these candidates, it is imperative to identify the supporting context that logically connects the query to the specific answer.
We define this pivotal bridging context as the \textit{clue}.
Figure~\ref{fig:EXAns2} depicts a representative scenario comprising a user query, candidate entities, and the associated clue sentence.
As observed, although the context contains multiple person entities, the correct answer can be identified by a clue sentence that semantically aligns with the intent of the query.
Moreover, these clue sentences exhibit significantly higher semantic similarity to the query compared to irrelevant background text.

Consequently, we extract potential clues by evaluating the semantic affinity between the query and context sentences.
To construct a comprehensive candidate clue set $\mathbb{C}_{\text{clue}}$ that retains all potential evidence while filtering out obvious noise, we employ a threshold-based selection strategy rather than a fixed-size truncation. 
Formally, sentences with a semantic similarity score exceeding a predefined threshold $T$ are aggregated into the set:
\begin{equation}\label{clueset}
	\mathbb{C}_{\text{clue}} = \left\{ s_j \in x_{\text{s}} \mid {\sigma}(s_j, x_{\text{q}}) \geq T \right\}
\end{equation}
This candidate clue set effectively highlights pertinent content, providing the necessary semantic context to disambiguate the entities in $\mathbb{C}_{\text{ans}}$.
With the candidate answer set $\mathbb{C}_{\text{ans}}$ and the candidate clue set $\mathbb{C}_{\text{clue}}$ successfully constructed, we are now positioned to evaluate their effectiveness.

\begin{table}[t]
	\centering
	\caption{Average results of the first compression stage.}
	\label{tab:compression_stats}
	\begin{tabular}{l c c c}
		\toprule
		Dataset & \multicolumn{1}{c}{$len(\hat{x})$} & \multicolumn{1}{c}{$\tau$} & $len(\mathbb{C}_{\text{ans}}) / len(\hat{x})$\\
		\midrule
		2wikimqa & 5020 & 33.78\% & 91.44\%\\
		narrativeqa & 5384 & 82.97\% & 90.71\%\\
		musique & 9025 & 45.64\% & 92.53\%\\
		\bottomrule
	\end{tabular}
    \vspace{-0.4cm}
\end{table}

\subsection{Motivation}\label{sec:motivation}

We conduct preliminary experiments of the compression method based on the two extracted sets.
Specifically, we form a compressed context by amalgamating all sentences from the candidate answer set $\mathbb{C}_{\text{ans}}$ and a subset of the candidate clue set $\mathbb{C}_{\text{clue}}$.
For the experimental configuration, we utilize the lightweight \texttt{en\_core\_web\_sm} model (occupies $\approx$ 12 MB for model weights) from the spaCy library~\cite{spacy2015industrial} as the entity classifier $M_{\text{cls}}$, and employ the \texttt{MiniLM-L6-v2} model (occupies $\approx$ 22 MB for model weights)~\cite{wang2020minilm} as the embedding model $M_{\text{emb}}$.
Regarding the clue set, we establish a semantic similarity threshold $T=0.5$, which serves as an empirical baseline to filter out obvious noise \cite{reimers2019sentence}.
In addition, we select the top-$K$ sentences with the highest similarity as the experimental clue set to avoid introducing excessive noise.
We evaluate the performance of our compression method using three representative LLMs with different context window lengths, \ie, \texttt{LLaMA2-7B-chat-4k} \cite{touvron2023llama}, \texttt{Longchat1.5-7B-32k} \cite{bai2024longbench}, and \texttt{Qwen2.5-7B-Instruct}~\cite{yang2024qwen2}, across three LongBench datasets \cite{bai2024longbench}, \ie, 2wikimqa, narrativeqa, and musique.
Figure~\ref{fig:PreEX} summarizes the performance variations with different $K$ values, while Table~\ref{tab:compression_stats} details for each dataset the average prompt length, the true compression ratio $\tau$, and the proportion of candidate answer sentences within the compressed prompt.

As illustrated in Figure~\ref{fig:PreEX} and Table \ref{tab:compression_stats}, the compression strategy demonstrates consistent effectiveness across different LLMs.
Specifically, Figure~\ref{fig:PreEX}(a) shows that despite achieving a significant compression ratio of nearly 83\% on the NarrativeQA dataset \cite{kocisky-etal-2018-narrativeqa}, the performance degradation with the \texttt{LLaMA2-7B-chat-4k} model remains within a marginal range of 2.41\% to 5.85\%.
Similar trends are observed for \texttt{Qwen2.5-7B-Instruct} and \texttt{Longchat1.5-7B-32k}, where the performance drops are 6.49\%--22.50\% and 17.64\%--21.12\%, respectively, corroborating that compression-induced degradation tends to be smaller for models with shorter context window lengths.
More compellingly, Figure~\ref{fig:PreEX}(a) and Figure~\ref{fig:PreEX}(b) reveal that on the 2WikiMQA dataset \cite{sugawara2020constructing}, prompt compression can even yield performance gains by reducing noise for certain models.
Specifically, with the \texttt{LLaMA2-7B-chat-4k} model, the F1-Score improves from 19.37 to 22.36 at $K = 7$, representing a enhancement of 3.00\%, while \texttt{Longchat1.5-7B-32k} achieves marginal improvements up to 15.44\% at $K = 3$.
Despite slight performance degradation observed for \texttt{Qwen2.5-7B-Instruct}, the overall effectiveness of the compression strategy remains noteworthy.
These results demonstrate that the compression strategy based on clues and answers achieves notable effectiveness across diverse model architectures even without an SLM.

Although these results are encouraging, distilling these candidate sentences into a concise context within a rigid token budget still presents challenges.
First, determining the optimal clue subset is non-trivial.
Simple heuristics of linearly expanding or shrinking the subset fail to consistently yield optimal results across different models and datasets.
For instance, as shown in Figure~\ref{fig:PreEX}(a), on the MuSiQue dataset using \texttt{LLaMA2-7B-chat-4k}, performance peaks at 9.48 with 7 clues but drops to 9.24 with 5 clues and 9.09 with 9 clues.
Similarly, Figure~\ref{fig:PreEX}(a) reveals that for \texttt{Longchat1.5-7B-32k} on the 2WikiMQA dataset, performance fluctuates non-monotonically, achieving 22.36 at $K = 3$, declining to 20.30 at $K = 7$, and subsequently rising to 23.20 at $K = 9$.
A comparable pattern is observed in Figure~\ref{fig:PreEX}(c) for \texttt{Qwen2.5-7B-Instruct} on MuSiQue, where performance initially decreases from 30.37 (uncompressed) to 27.75 at $K = 3$, then progressively recovers to 30.17 at $K = 9$.
These observations collectively highlight that the decision of subset size should not be made in isolation but must strictly account for the candidate answer set, as the utility of a retrieved clue is fundamentally contingent upon its capacity to bridge the query to specific answers.
Second, according to results in Table \ref{tab:compression_stats}, the candidate answer set remains excessively large, dominating over 90\% of the compressed context. 
Because mere retention of all answers leads to uncontrollable compression ratios, it is imperative to filter irrelevant answers based on the spatial support provided by high-confidence evidence in the clue set.
This necessitates a synergistic refining strategy ensuring clues efficiently support answers while answers are rigorously validated by their corresponding clues.

To overcome the limitations of current compression method, we further refine the extracted sets. 
Considering the difficulty of determining the clue subset size, the refining process employs an orthogonal residual retrieval strategy that adaptively decides $K$ based on coverage of the candidate answer set.
Moreover, to mitigate the issue of excessive candidate answers, we also incorporates a dedicated importance metric that leverages the spatial proximity of clues, which ensures a highly coherent and evidence-grounded answer subset.  
The initial compression method and the refining process together form the fundamental framework of \method, which will be detailed in the next section.
	
	\section{System Design}\label{sec:algorithm}
	\begin{figure*}[t]
\centering
\includegraphics[height=7.5cm]{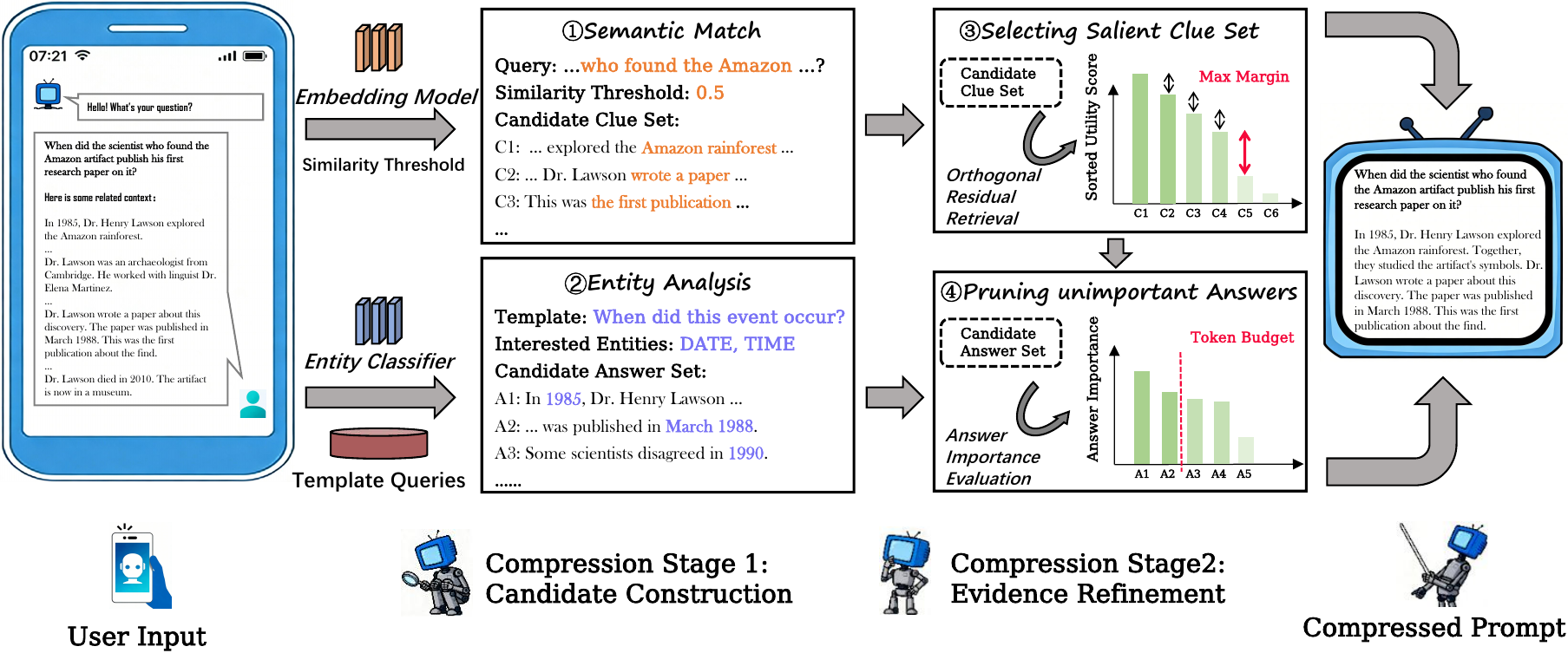}
\caption{Illustration of the compression process of \method.}
\label{fig:overview}
\end{figure*}

\subsection{Overview of \method}
The process of \method is illustrated in Figure~\ref{fig:overview}.
In the candidate filtering stage, \method creates a clue set by calculating the semantic similarity between the user query and all context sentences (Step~\circled{1}).
Simultaneously, it analyzes the query intent to recognize interested entities and filters the context into an answer set of sentences containing these entities (Step~\circled{2}).
Subsequently, the workflow transitions to the evidence refining stage.
This stage features an iterative orthogonal retrieval strategy to distill a salient clue subset (Step~\circled{3}), paired with a pruning mechanism that filters answers based on their relevance to both the query and the retrieved clues (Step~\circled{4}).
Finally, the refined clue set and the pruned answer set are merged to form the compressed context.
A more detailed description is provided in Algorithm~\ref{alg1}.

\subsection{Clue Refining}
In Section~\ref{sec:motivations}, the candidate filtering stage relies simply on similarity ranking to establish the candidate clue set $\mathbb{C}_{\text{clue}}$. 
However, complex queries often encompass multiple semantic dimensions \cite{ZHU2024111301}, such as temporal constraints, specific entities, and logical relationships, a naive selection of top-ranked sentences tends to result in redundancy.
Such approaches disproportionately retrieves sentences aligning with the dominant semantic aspect, thereby overlooking subtle yet critical details.
To address this limitation and construct a compact yet comprehensive clue subset $\mathbb{\hat{C}}_{\text{clue}}$, we employ an iterative retrieval strategy grounded in orthogonal vector decomposition~\cite{pati1993orthogonal}.
The strategy operates by sequentially identifying the unexplained semantic residuals of the query.
By selecting clues that align with these residuals, the algorithm ensures that each addition contributes unique information, thereby maximizing semantic coverage with minimal redundancy.

Let $q^{(t)}_{\text{res}}$ denote the query residual vector at iteration $t$, which represents the semantic information of the query that has not yet been explained by the selected clues.
The vector is initialized as $x_{\text{q}}$.
In each iteration, we calculate the utility score for each remaining clue $c_i$ based on its semantic alignment with the current residual, \ie, $\sigma(c_i, q_{\text{res}}^{(t)})$, and the clue $c^*_t$ with the highest score is selected.
Subsequently, to encourage diversity in the next iteration, we perform a soft orthogonalization update to the query residual.
This step conceptually subtracts the semantic direction of $c^*_t$ from the query.
The update rule is defined as:
\begin{equation}\label{soft_orth}
	q_{\text{res}}^{(t+1)} = q_{\text{res}}^{(t)} - \mu_t \cdot \frac{q_{\text{res}}^{(t)} \cdot c^*_t}{\|c^*_t\|^2} c^*_t
\end{equation}
where $\mu_t \in (0, 1]$ is a dynamic penalty coefficient that controls the aggressiveness of the subtraction.

\begin{algorithm}[t]
	\caption{Compression process of \method.}
	\label{alg1}
	\KwIn{User query $x_{\text{q}}$, Context $x_{\text{s}} = \{s_1, s_2, \dots\}$, Embedding model $M_{\text{sem}}$, Entity classifier $M_{\text{cls}}$, Template queries $\{q_1,q_2, \dots\}$, Token budget $\mathcal{B}$, Entity sets $\{\mathbb{E}_1, \mathbb{E}_2,\dots\}$, Similarity threshold $T$}
	\KwOut{Compressed prompt $\hat{x}$.}
	Create the clue set $\mathbb{C}_{\text{clue}}$ based on Eq.~\eqref{clueset}\;
	Create the interested entity set $\hat{\mathbb{E}}$ based on Eq.~\eqref{sim}\eqref{Ien}\;
	Create the answer set $\mathbb{C}_{\text{ans}}$ based on Eqs.~\eqref{ansset}\;
	$t=0, ~q^{(t)}_{\text{res}} = x_{\text{q}}$\;
	\Repeat{${\mathbb{C}}_{\text{clue}} = \emptyset$}{
		$c^*_t = \text{argmax}_{c_i \in {\mathbb{C}}_{\text{ans}}}\sigma(c_i, q_{\text{res}}^{(t)})$\;
		Update $q^{(t)}_{\text{res}}$ based on Eqs.~\eqref{soft_orth}\eqref{mu}\;
		${\mathbb{C}}_{\text{clue}} = {\mathbb{C}}_{\text{clue}} \setminus \{c^*_t\}, t=t+1$\;
	}
	Decide the size of the clue set $K$ based on Eq.~\eqref{elbow}\;
	Create the clue subset $\hat{\mathbb{C}}_{\text{clue}} = \{c^*_t \mid t \leq K\}$\;
	\For{$a_i \in \mathbb{C}_{\text{ans}}$}{
		\For{$e_j \in \{e_j \mid e_j \in M_{\text{cls}}(a_i) \cup e_j \in \hat{E}\}$}{
			Calculate $I_{e_j}$ based on Eq. \eqref{imp_entity}\;
		}
		Calculate the answer importance $I_{a_{i}} = \sum_j I_{e_j}$\;
	}
	$\mathcal{B}_{\text{ans}} = \mathcal{B} - len(x_{\text{q}}) - \sum_{s_i \in \hat{\mathbb{C}}_{\text{clue}}} n^i$\;
	Create the answer subset $\hat{\mathbb{C}}_{\text{ans}}$ by selecting most important answers with in Budget $\mathcal{B}_{\text{ans}}$\;
	$\hat{x} = (x_{\text{q}}, \hat{\mathbb{C}}_{\text{ans}}, \hat{\mathbb{C}}_{\text{clue}})$\;
\end{algorithm}

Crucially, to account for the intrinsic dependency between the candidate sets, $\mu_t$ is rigorously calibrated based on the clue’s capacity to bridge the query to interested entities within candidate answers.
A higher $\mu_t$ implies that the clue already supports a significant portion of candidate answers. In such cases, the semantic dimension represented by this clue is already well-covered by the current answer set, so we apply a more aggressive subtraction to encourage the next iteration to explore alternative semantic directions, thereby preventing redundancy.
To capture the local contextual information around the selected clue, we consider its neighboring sentences in the original context. Let $s_{pre}$ and $s_{succ}$ denote the nearest preceding and succeeding sentence of $c^*_t$, respectively.
We calculate $\mu_t$ as the ratio of unique answer entities covered by this local context window to the total unique entities in the candidate answer set $\mathbb{C}_{\text{ans}}$:
\begin{equation}\label{mu}
	\mu_t = \frac{|\mathbb{V}(c^*_t) \cup \mathbb{V}(s_{pre}) \cup \mathbb{V}(s_{succ})|}{|\mathbb{V}({\mathbb{C}_{\text{ans}}})|}
\end{equation}
where $\mathbb{V}(\cdot)$ denotes the set of unique answer entities.

This iterative process generates a sequence of clues ranked by their marginal utility.
To automatically determine the size of $\mathbb{\hat{C}}_{\text{clue}}$, we analyze the sequence of maximum scores using the elbow method \cite{bholowalia2014ebk}.
The stopping point $K$ is identified as the point of maximum curvature in the score curve, which indicates the iteration where the marginal gain of adding a new clue begins to diminish significantly.
Mathematically, this is equivalent to finding the index that maximizes the second-order difference of the score sequence:
\begin{align}\label{elbow}
    K = \text{argmax}_{k} \big[ & \sigma(c^*_{k-1}, q_{\text{res}}^{(k-1)}) \nonumber \\
    & - 2\sigma(c^*_{k}, q_{\text{res}}^{(k)}) + \sigma(c^*_{k+1}, q_{\text{res}}^{(k+1)}) \big]
\end{align}
By selecting the top-$K$ clues, we ensure a balance between semantic coverage and token consumption.

\subsection{Answer Pruning}
To address the issue of excessive candidate answers revealed in Section~\ref{sec:motivation}, we proceed to evaluate the importance of each sentence within the candidate answer set $\mathbb{C}_{\text{ans}}$ following the derivation of the clue subset $\hat{\mathbb{C}}_{\text{clue}}$.
Our evaluation metric is grounded in the intuition of contextual evidence propagation \cite{mihalcea2004textrank}, where high-confidence clues are semantic anchors that radiate relevance to spatially adjacent entities.

For a unique entity $e_j$, considering all its occurrences $\{e^1_{j}, e^2_{j}, \dots\}$ in the text, we calculate its global importance score $I_{e_j}$ by identifying its most supported instance:
\begin{equation}\label{imp_entity}
	I_{e_j} = \max_{m} \left( \sum_{c_k \in \hat{\mathbb{C}}_{\text{clue}}} \sigma(c_k, x_{\text{q}}) \cdot \exp\left(-\frac{\text{D}(e^m_{j}, c_k)^2}{2\eta^2}\right) \right)
\end{equation}
where $\text{D}(e^m_{j}, c_k)$ denotes the number of tokens separating the entity occurrence $e^m_{j}$ from the clue $c_k$ in the original context, and the exponential term enforces a spatial constraint regulated by a bandwidth parameter $\eta$.
This interaction ensures that an entity receives a high score only if it is situated in the immediate vicinity of a relevant clue.

We define the bandwidth parameter $\eta$ as the mean token count of the sentences within the clue subset $\hat{\mathbb{C}}_{\text{clue}}$:
\begin{equation}\label{eta}
	\eta = \sum_{c \in \hat{\mathbb{C}}_{\text{clue}}} \text{len}(c) / {|\hat{\mathbb{C}}_{\text{clue}}|}
\end{equation}
The rationale is that the effective scope of a clue's influence should reflect the granularity of the context: longer sentences imply broader semantic units, so the relevance propagation should extend farther to capture meaningful contextual relationships.

For an answer sentence $a_i$ containing a set of entities $\{e_j \mid e_j \in M_{\text{cls}}(a_i)\}$, its overall importance is computed as the sum of its constituent entity scores:
\begin{equation}
\label{imp_answer}
    I_{a_i} = \sum_{e_j \in M_{\text{cls}}(a_i)} I_{e_j}
\end{equation}
This metric discriminates between contextually grounded answers and irrelevant mentions, facilitating the precise pruning of the answer set to strictly adhere to the target token budget.
	
	\section{Experiments and Evaluation}\label{sec:evaluation}
	\subsection{Experiment Platform}
To thoroughly evaluate the performance of \method against baseline methods in realistic scenarios, we simulate two prevalent deployment modes for edge-side intelligent agents. 
The detailed technical specifications for the hardware platforms used in these modes are listed in Table~\ref{tab:device}. 
The two modes are defined as follows:
\begin{itemize}
	\item \textbf{Edge Node with Local Inference:} This mode represents scenarios with robust local processing capabilities, \eg, autonomous vehicles and industrial robots. 
	We employ an NVIDIA Jetson AGX Orin as the edge device. 
	In this setup, both the prompt compression via \method and the LLM inference are executed entirely on the local node.
	\item \textbf{Edge Node with Cloud Offloading:} This mode simulates environments with weak computational capabilities where onboard LLM inference is impractical. We utilize a Huawei Nova 12 smartphone as the mobile device. Here, prompt compression is performed locally to reduce data size, while the final generation request is offloaded to a cloud LLM API.
\end{itemize}

\begin{figure}[h]
	\centering
	\includegraphics[height=2.5cm]{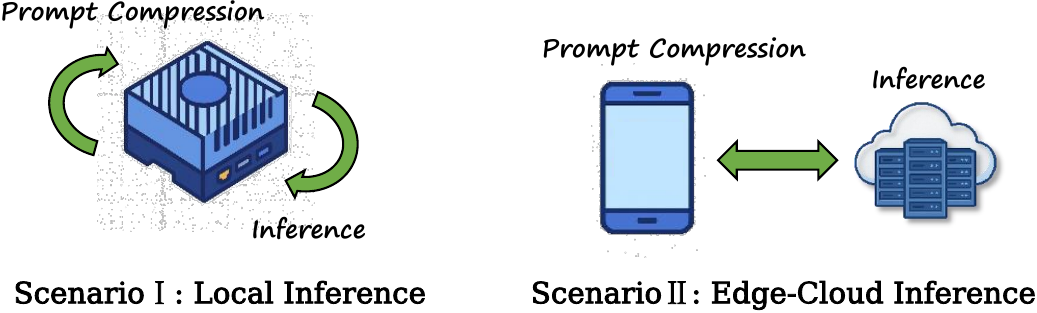}
	\caption{Different experimental scenarios.}
	\centering
	\label{fig:scenarios}
\end{figure}

The underlying software platform for these experiments is built upon the PyTorch Deep Learning library~\cite{paszke2019pytorch}. 
For the SLMs or other models used by baselines and \method during the compression, we deploy the models using the Transformers framework \cite{wolf2020transformers}. 
For the local inference LLMs, we base our model deployment on the vLLM framework \cite{kwon2025vllm}, a library that facilitates the implementation of LLM inference acceleration techniques.
Furthermore, to address potential inconsistencies arising from the distinct operating systems and hardware configurations of the NVIDIA Jetson and the mobile device, we standardized the execution environment using a Docker containerization framework~\cite{docker1}.
The semantic similarity threshold $T$ and the selections for models $M_{\text{cls}}$ and $M_{\text{emb}}$ remain consistent with the settings of the preliminary experiments described earlier in Section 3.
The two scenarios are illustrated in Figure~\ref{fig:scenarios}.

\subsection{Setup of Experiments}

\begin{table}[t]
	\caption{Edge device technical specifications.}
	\centering
	\resizebox{1.\columnwidth}{!}{%
		\begin{tabular}{l|cc}
			\toprule
			\textbf{}  & \textbf{Jetson AGX Orin} & \textbf{Huawei Nova 12} \\
			\midrule
			\textbf{AI performance}   & 200 TOPS      & 12 TOPS   \\
			\midrule
			\textbf{CPU Type} &   8-core Cortex ARM 8       & Snapdragon 778G 4G \\
			\midrule
			\textbf{GPU Type}   & 1792-core Volta      & Adreno 642L   \\
			\midrule
			\textbf{ROM} & 64 GB LPDDR5  & 8 GB LPDDR4x \\
			\bottomrule
		\end{tabular}%
	}
	\label{tab:device}
    \vspace{-0.4cm}
\end{table}

\textbf{Datasets.}
To comprehensively evaluate the capabilities of \method in handling diverse long-context QA scenarios, we utilize six representative datasets from the LongBench benchmark.
Each dataset contains 200 questions, categorized into three distinct challenging clusters based on their primary challenges:

\textbf{\textit{1) Complex Reasoning.}}
This category assesses the model's ability to perform multi-hop reasoning and integrate dispersed information from multiple sources.
MuSiQue~\cite{trivedi2022musique} and HotpotQA~\cite{yang2018hotpotqa} challenge the model to aggregate evidence across connected logical steps or multiple documents, specifically designed to prevent naive disconnected reasoning shortcuts.
Additionally, 2WikiMQA~\cite{sugawara2020constructing} introduces hybrid complexity by combining unstructured text with structured knowledge, requiring intricate specific logical paths.

\textbf{\textit{2) Technical Synthesis.}}
These datasets assess the model's adaptability to specialized document structures and domain-specific terminologies. 
Qasper~\cite{dasigi2021dataset} simulates academic literature reviews, requiring evidence synthesis from full NLP papers.
MultiFieldQA~\cite{bai2024longbench} broadens this scope to interdisciplinary information needs, curating diverse sources such as legal and government reports to test robustness across approximately 10 different fields.

\textbf{\textit{3) Narrative Comprehension.}}
This category focuses on inferring implicit answers from lengthy movie scripts and literary works. Represented by NarrativeQA~\cite{kocisky-etal-2018-narrativeqa}, it demands a holistic understanding of complex plot developments, character relationships, and thematic elements that span across the entire document.

In addition to these benchmarks, we incorporate a real-world dataset in open-domain environments to test the efficiency of compression methods in filtering non-content artifacts while preserving essential evidence:

\textbf{\textit{NaturalQuestions}}~\cite{kwiatkowski2019natural} serves as a testbed for evaluating the compressor's capability to pinpoint key information within real-world documents.
It provides raw HTML contexts laden with structural noise, \eg, tables and tags.
The dataset is preprocessed using the BeautifulSoup library~\cite{richardson2024beautiful}.

\begin{table*}[t]
	\centering
	\caption{7B models' F1-Score performance of different methods under a 2000-token budget on LongBench.}
	\resizebox{1.9\columnwidth}{!}{
		\begin{tabular}{c|c|ccccccc}
			\toprule
			\multirow{2}{*}{\textbf{Models}} & \multirow{2}{*}{\textbf{Methods}} &  \multicolumn{7}{@{}c}{{\bf LongBench}} \\
			& & Qasper & MultiFieldQA & NarrativeQA & MuSiQue & HotpotQA & 2WikiMQA & AVG \\
			\midrule
			\midrule
			\multirow{5}{*}{\texttt{LLaMA2-7B-chat-4k}} & Original Prompt & 21.90 & 36.63 & 18.64 & 7.65 & 27.28 & 31.96 & 24.01 \\
			\cmidrule (r){2-2}\cmidrule (lr){3-9}
			& LLMLingua & 14.86 & 24.32 & 11.43 & 6.59 & 24.55 & 20.28 & 17.01 \\
			& {LLMLingua-2} &  \textbf{24.17} & 31.90 & 5.85 & 11.18 & 25.72 & 28.52 & 21.22 \\
			& {LongLLMLingua} &  18.54 & 28.00 & 12.95 & 15.66 & 28.81 & \textbf{31.96} & 22.65 \\
			\cmidrule (r){2-2}\cmidrule (lr){3-9}
			& {\cellcolor[rgb]{0.925,0.957,1}}\textbf{\method (Ours)} & {\cellcolor[rgb]{0.925,0.957,1}}23.98 & {\cellcolor[rgb]{0.925,0.957,1}}\textbf{40.30} & {\cellcolor[rgb]{0.925,0.957,1}}\textbf{19.54} & {\cellcolor[rgb]{0.925,0.957,1}}\textbf{16.36} & {\cellcolor[rgb]{0.925,0.957,1}}\textbf{35.17} & {\cellcolor[rgb]{0.925,0.957,1}}29.99 & {\cellcolor[rgb]{0.925,0.957,1}}\textbf{27.56} \\
			\midrule
			\midrule
			\multirow{5}{*}{\texttt{Longchat1.5-7B-32k}} & Original Prompt & 28.17 & 41.72 & 20.07 & 9.46 & 31.10 & 19.37 & 24.98  \\
			\cmidrule (r){2-2}\cmidrule (lr){3-9}
			& LLMLingua & 16.59 & 21.39 & 10.44 & 7.05 & 21.89 & 18.82 & 16.03  \\
			& {LLMLingua-2} &  26.66 & 32.29 & 5.34 & 10.36 & 25.81 & 23.67 & 20.69 \\
			& {LongLLMLingua} &  16.82 & 25.60 & 10.08 & 14.20 & 25.44 & \textbf{26.55} & 19.78 \\
			\cmidrule (r){2-2}\cmidrule (lr){3-9}
			& {\cellcolor[rgb]{0.925,0.957,1}}\textbf{\method (Ours)} & {\cellcolor[rgb]{0.925,0.957,1}}\textbf{28.99} & {\cellcolor[rgb]{0.925,0.957,1}}\textbf{41.73} & {\cellcolor[rgb]{0.925,0.957,1}}\textbf{18.41} & {\cellcolor[rgb]{0.925,0.957,1}}\textbf{18.61} & {\cellcolor[rgb]{0.925,0.957,1}}\textbf{34.21} & {\cellcolor[rgb]{0.925,0.957,1}}25.61 & {\cellcolor[rgb]{0.925,0.957,1}}\textbf{27.93} \\
			\midrule
			\midrule
			\multirow{5}{*}{\texttt{Qwen2.5-7B-Instruct}} & Original Prompt & 43.67 & 52.76 & 28.84 & 30.37 & 57.84 & 47.13 & 43.44 \\
			\cmidrule (r){2-2}\cmidrule (lr){3-9}
			& LLMLingua & 21.00 & 22.25 & 8.66 & 13.36 & 27.85 & 27.62 & 20.12 \\
			& {LLMLingua-2} &  \textbf{42.20} & 39.55 & 12.21 & 13.87 & 35.06 & 37.12 & 30.00 \\
			& {LongLLMLingua} &  27.00 & 30.41 & 12.27 & 20.60 & 38.79 & \textbf{41.96} & 28.51 \\
			\cmidrule (r){2-2}\cmidrule (lr){3-9}
			& {\cellcolor[rgb]{0.925,0.957,1}}\textbf{\method (Ours)} & {\cellcolor[rgb]{0.925,0.957,1}}36.62 & {\cellcolor[rgb]{0.925,0.957,1}}\textbf{45.52} & {\cellcolor[rgb]{0.925,0.957,1}}\textbf{17.32} & {\cellcolor[rgb]{0.925,0.957,1}}\textbf{25.97} & {\cellcolor[rgb]{0.925,0.957,1}}\textbf{46.28} & {\cellcolor[rgb]{0.925,0.957,1}}39.81 & {\cellcolor[rgb]{0.925,0.957,1}}\textbf{35.25} \\
			\bottomrule
		\end{tabular}
	}
	\label{tab:main_result_long_context}
\end{table*}

\textbf{Models.}
Given that 7B parameter models represent the current industry standard for deployable LLMs on edge devices, we select three representative open-source 7B models for evaluation.
These models are specifically chosen to span a range of distinct context window sizes, enabling us to systematically investigate the impact of native context capacity on compression efficacy:

\textbf{\textit{1) LLaMA2-7B-chat-4k}}~\cite{touvron2023llama} is a widely adopted open-source foundation model series developed by Meta AI. It features a decoder-only transformer architecture with a context window of 4k tokens, serving general chat capabilities.

\textbf{\textit{2) LongChat-1.5-7B-32k}}~\cite{li2023long} addresses the long-context limitation of standard models. Developed by LMSYS based on the LLaMA architecture, it is fine-tuned to support context windows up to 32k tokens, making it ideal for evaluating performance on extensive retrieval results.

\textbf{\textit{3) Qwen2.5-7B-Instruct}}~\cite{yang2024qwen2} represents the state-of-the-art in the 7B class. Released by Alibaba Cloud, it features architectural optimizations and an expanded context length of 128k tokens, providing a strong baseline for modern LLMs.

Complementing these local models, we employ a commercial API to simulate cloud offloading scenarios:

\textbf{\textit{4) GPT-3.5-Turbo}}~\cite{kocon2023chatgpt} represents the commercial cloud-based inference API of OpenAI, allowing us to assess the compatibility and efficiency of \method in mobile-to-cloud architectures.
Specifically, we utilize the \texttt{gpt-3.5-turbo-16k} variant, which features a 16,384-token context window.

\textbf{Baselines.}
We compare \method against three state-of-the-art prompt compression algorithms:

\textbf{\textit{1) LLMLingua}}~\cite{jiang2023llmlingua} is a coarse-to-fine compression method based on information theory. 
It utilizes the \texttt{LLaMA2-7B-hf} model~\cite{touvron2023llama} (occupies $\approx$ 14 GB for model weights) to calculate token-level perplexity, and removes tokens with lower perplexity.

\textbf{\textit{2) LLMLingua-2}}~\cite{pan2024llmlingua} improves upon LLMLingua by replacing perplexity-based heuristics with a data-driven approach. 
It uses a BERT-based model \texttt{XLM-RoBERTa-Large}~\cite{conneau2020xlmr} (occupies $\approx$ 2.6 GB for model weights) to directly predict token importance.

\textbf{\textit{3) LongLLMLingua}}~\cite{jiang2024longllmlingua} extends LLMLingua with additional optimizations for long-context scenarios. 
In addition to using the \texttt{LLaMA2-7B-hf} model to calculate perplexity, it employs a document reordering mechanism to enhance key information retention at the beginning of the prompt.

\begin{figure}[t]
	\hspace*{0.2cm}
	\begin{subfigure}{0.20\textwidth}
		\centering
		\includegraphics[height=3.3cm]{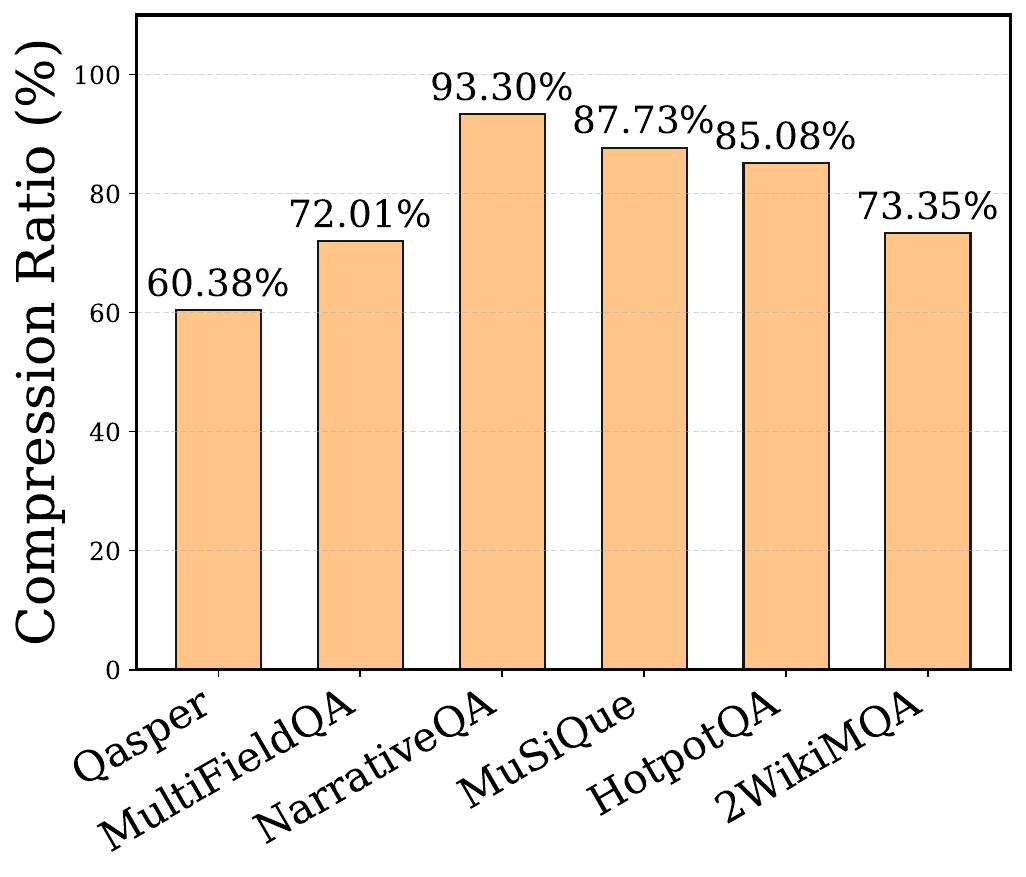}
		\caption{Compression ratios.}
		\label{fig:tau}
	\end{subfigure}
	\hspace{0.02\textwidth}
	\begin{subfigure}{0.20\textwidth}
		\centering
		\includegraphics[height=3.3cm]{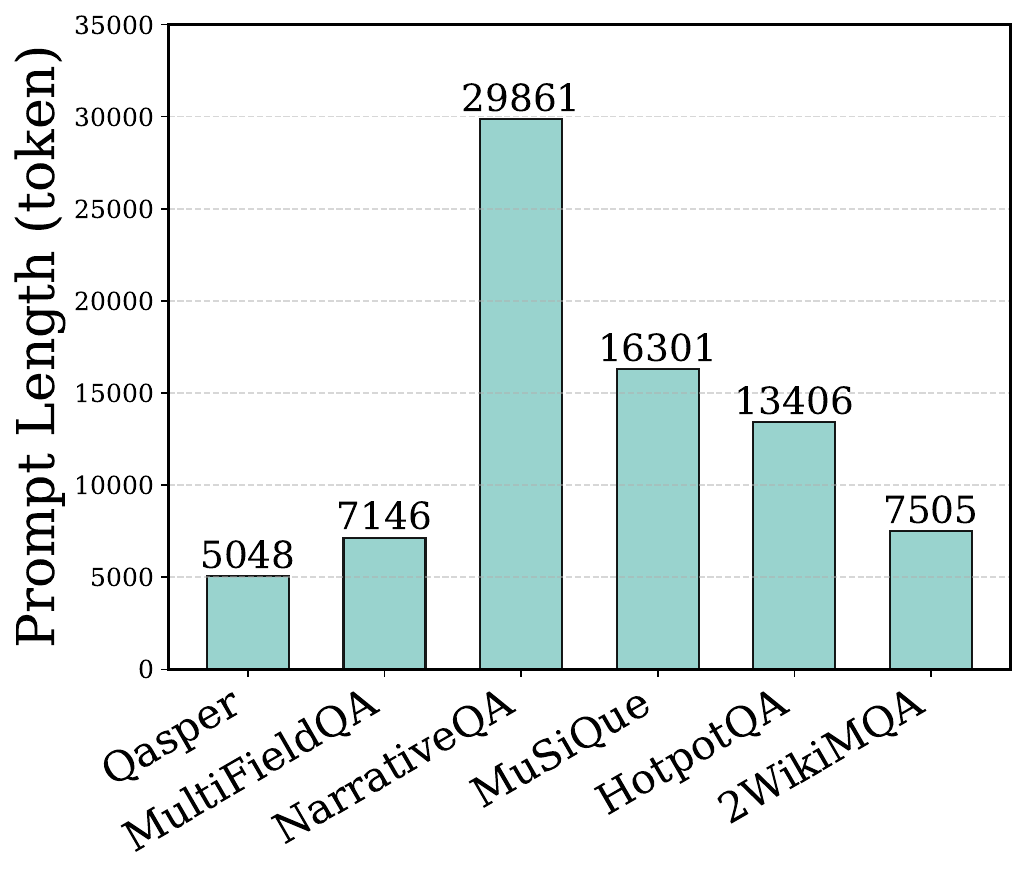}
		\caption{Prompt lengths.}
		\centering
		\label{fig:len}
	\end{subfigure}
	\caption{Dataset attributes on LongBench and the memory overhead of different compression methods.}
    \vspace{-0.4cm}
	\label{fig:M_L_tau}
\end{figure}

\begin{figure*}[t]
	\hspace*{-0.3cm}
	\begin{subfigure}{0.3\textwidth}
		\centering
		\includegraphics[height=4cm]{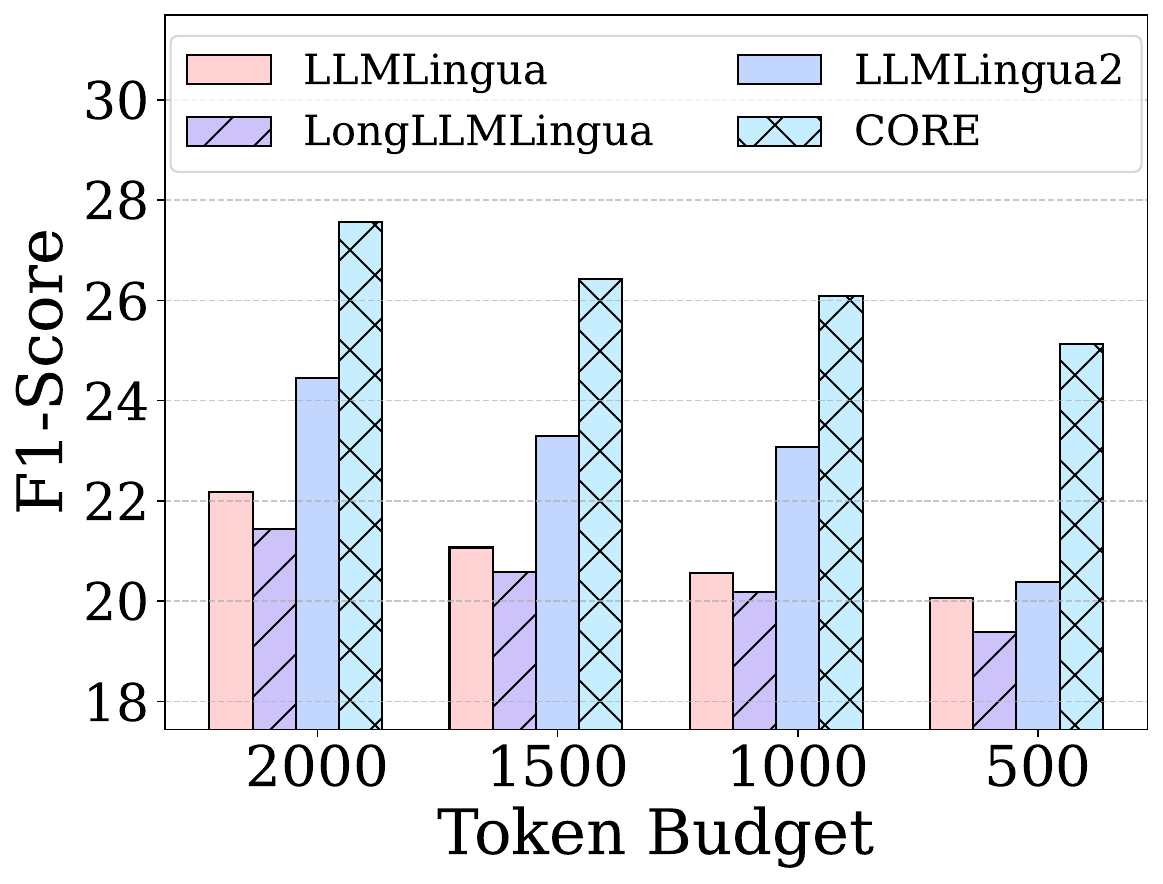}
		\caption{LLaMA2-7B-chat-4k}
		\centering
		\label{fig:llama2_7b_chat_score}
	\end{subfigure}
	\hspace{0.03\textwidth}
	\begin{subfigure}{0.3\textwidth}
		\centering
		\includegraphics[height=4cm]{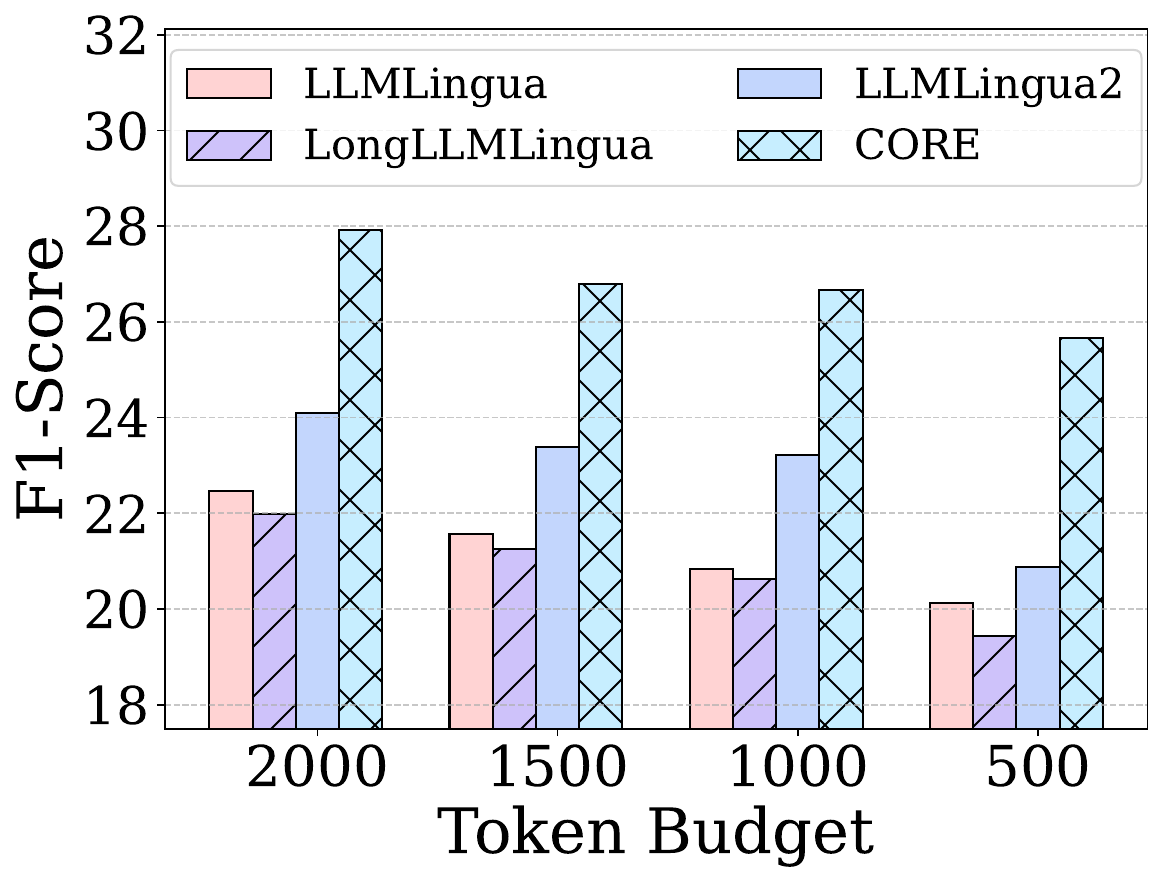}
		\caption{Longchat1.5-7B-32k}
		\label{fig:longchat_v1_5_7b_score}
	\end{subfigure}
	\hspace{0.03\textwidth}
	\begin{subfigure}{0.3\textwidth}
		\centering
		\includegraphics[height=4cm]{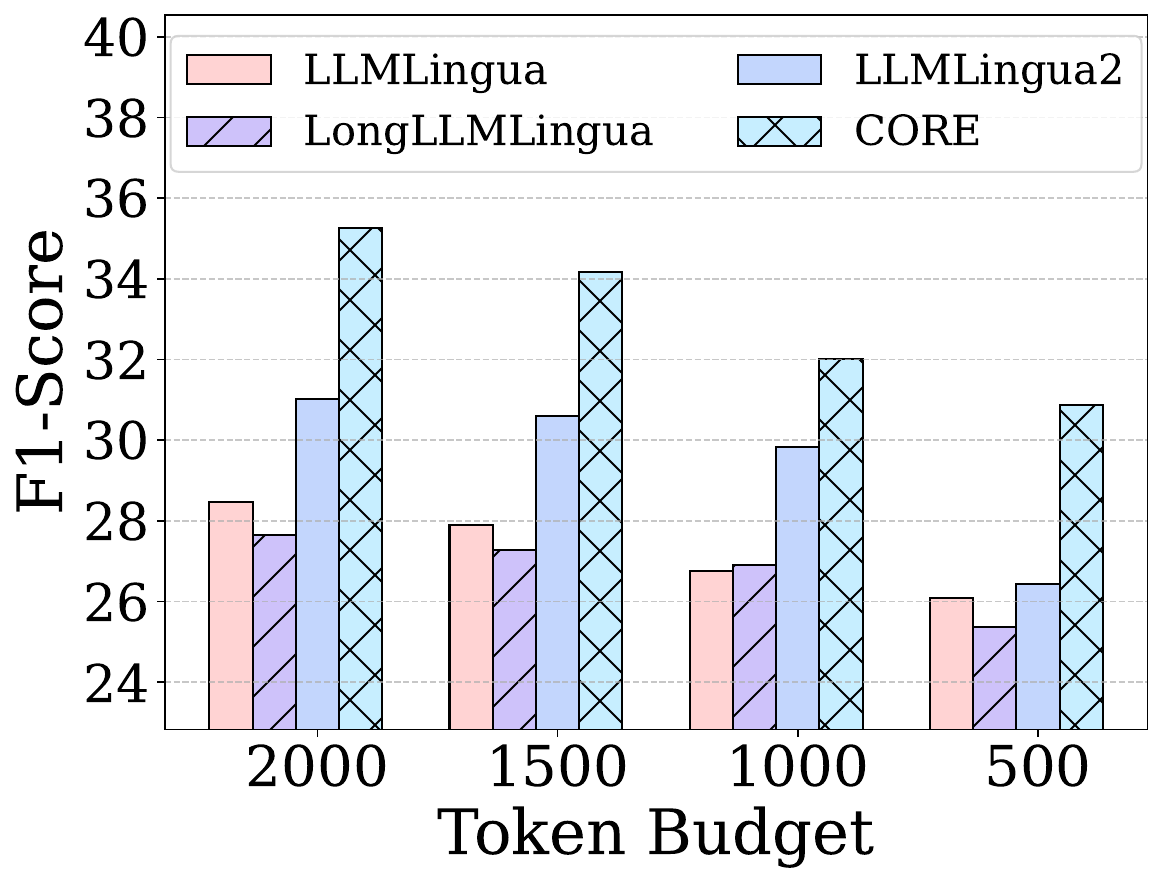}
		\caption{Qwen2.5-7B-Instruct}
		\label{fig:qwen2_5_7b_instruct_score}
	\end{subfigure}
	\caption{Average F1-Score variation with token budget across different models on the LongBench datasets.}
	\label{fig:Score}
\end{figure*}

\begin{figure*}[t]
	\hspace*{-0.3cm}
	\begin{subfigure}{0.31\textwidth}
		\centering
		\includegraphics[height=4cm]{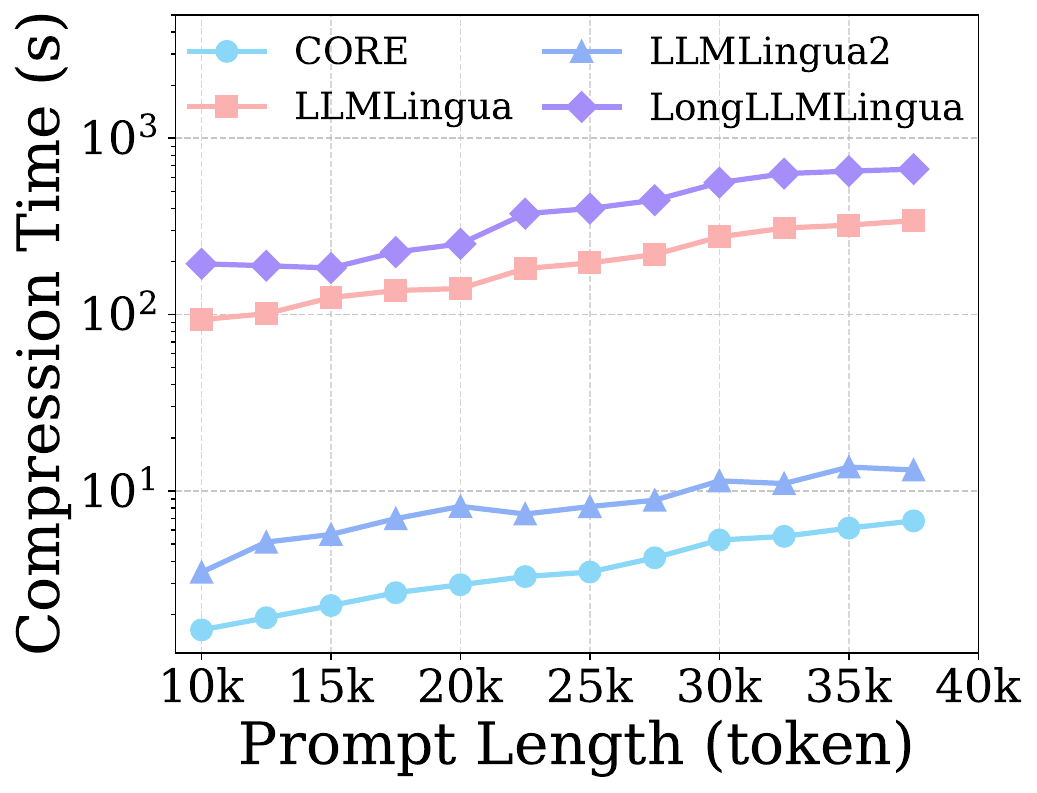}
		\caption{Compression Time}
		\centering
		\label{fig:EXtime}
	\end{subfigure}
    \hspace{0.03\textwidth}
    \begin{subfigure}{0.31\textwidth}
		\centering
		\includegraphics[height=4cm]{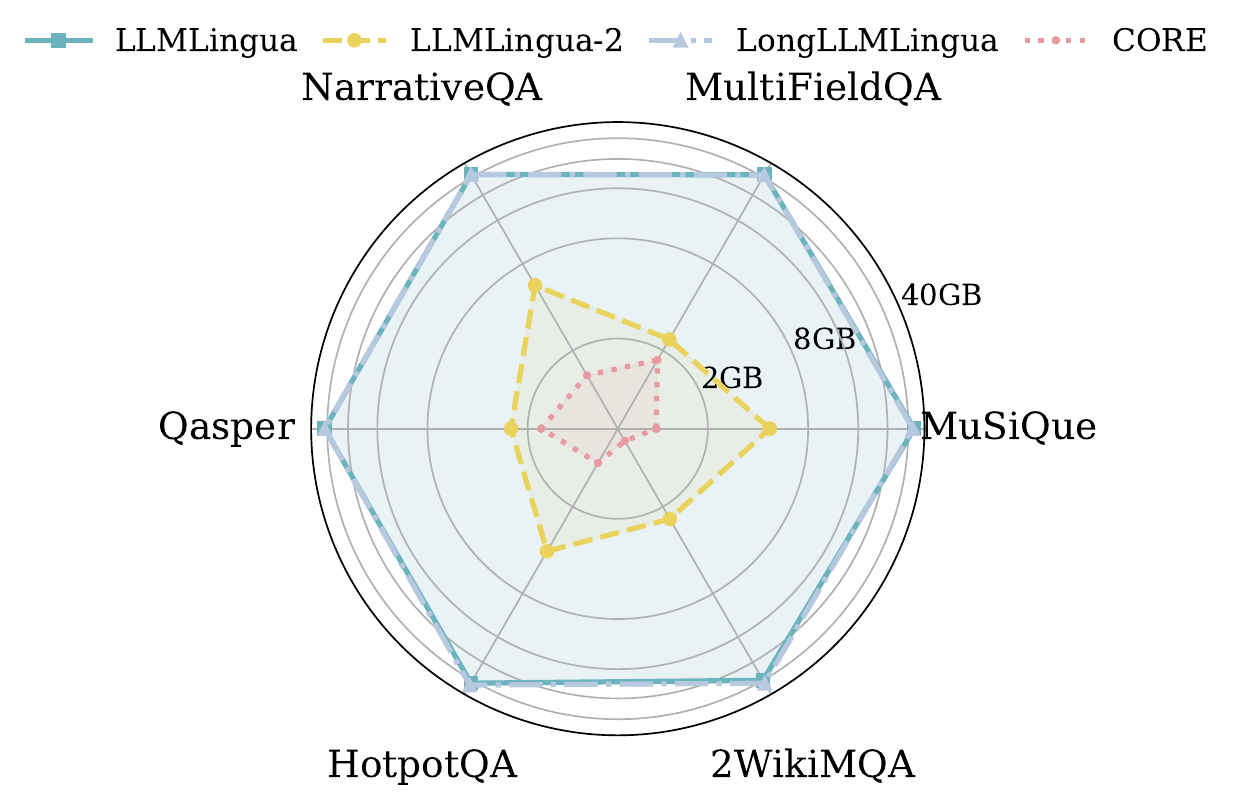}
		\caption{Memory overhead.}
		\centering
		\label{fig:Memory}
	\end{subfigure}
    \hspace{0.03\textwidth}
	\begin{subfigure}{0.31\textwidth}
		\centering
		\includegraphics[height=4cm]{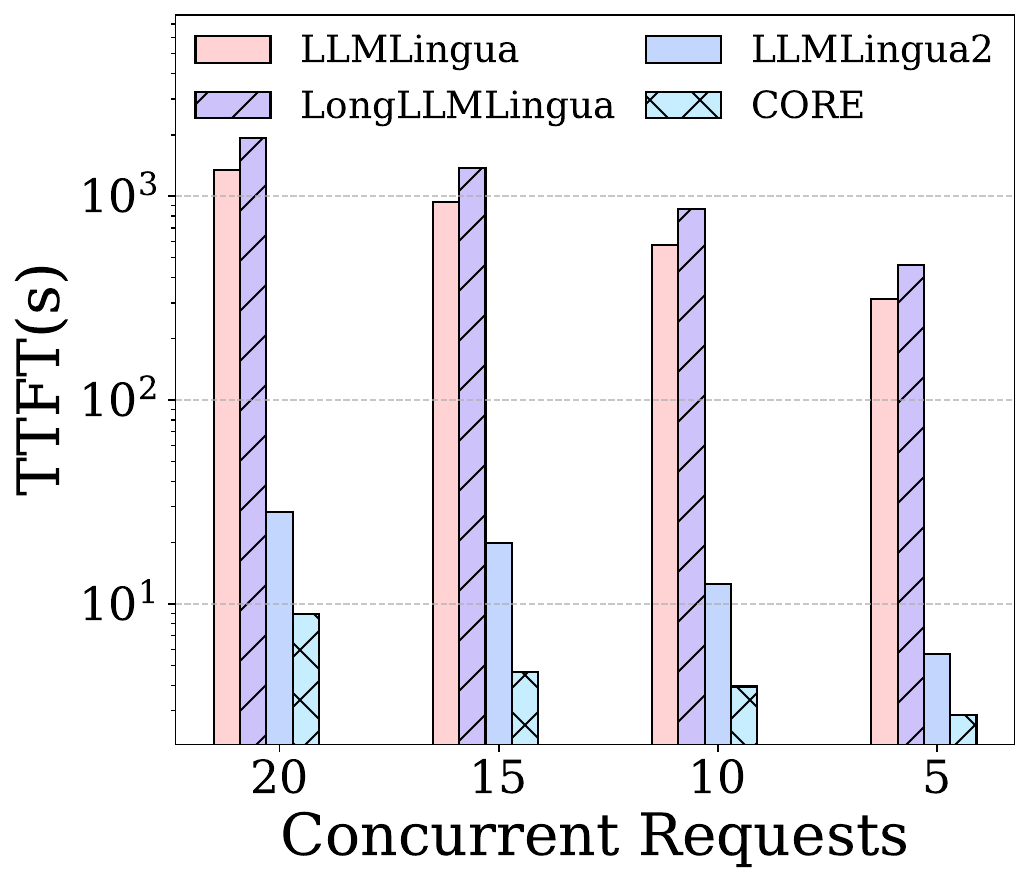}
		\caption{TTFT}
		\label{fig:EXTTFT}
	\end{subfigure}
	\caption{Latency analysis of different prompt compression approaches on LongBench datasets.}
	\label{fig:TTFT1}
    \vspace{-0.4cm}
\end{figure*}

\textbf{Metrics.}
We adopt five metrics to evaluate compression efficiency and inference quality:

\textbf{\textit{1) Compression Time}} measures the latency specifically incurred by the compression process, reflecting the computational efficiency of the compressor.

\textbf{\textit{2) Memory Overhead}} tracks the peak GPU memory usage during compression, a critical factor determining deployment feasibility on resource-constrained edge devices.

\textbf{\textit{3) Energy Consumption}} quantifies the total power usage during execution, calculated as the product of voltage, current, and time.
Battery Guru software is used to measure the average voltage and current during execution.

\textbf{\textit{4) Time To First Token (TTFT)}} measures the latency from sending the request to receiving the first token. A lower TTFT indicates reduced input processing load.


\textbf{\textit{5) F1-Score}} is computed as the harmonic mean of token-level precision and recall. 
Precision measures the proportion of overlapping tokens in the response, while recall measures the proportion of overlapping tokens in the reference answer.

\subsection{Edge-Only Experiment}
In this section, we conduct experiments with the local inference edge node scene on the LongBench datasets.

\textbf{Main Results.}
We first evaluate the F1-Score performance of \method against baselines, and the results are detailed in Table~\ref{tab:main_result_long_context}.
Our method demonstrates superior efficacy across all models.
Specifically, it achieves an average F1-score of 27.56 with the \texttt{LLaMA2-7B-chat-4k} model, marking a substantial improvement of 29.88\% over LLMLingua-2 and 21.68\% over LongLLMLingua.
With the model \texttt{Qwen2.5-7B-Instruct}, \method remains the most robust strategy with an average score of 35.25, surpassing the strongest baseline LLMLingua-2 by 17.50\%. 
Furthermore, \method even surpasses the uncompressed baseline by 11.81\% with the \texttt{Longchat1.5-7B-32k} model, underscoring its capacity.
Beyond average performance, \method exhibits exceptional robustness on specific datasets requiring complex information synthesis.
Based on the compression ratio results in Figure~\ref{fig:len_tau}, in the case of reasoning-intensive HotpotQA, \method achieves a score of 35.17 with the \texttt{LLaMA2-7B-chat-4k} model, significantly outperforming the original prompt by 28.92\% while keeping a compression ratio of 85.08\%.
This indicates that \method well preserves critical logical chains, thereby clarifying the reasoning path for the model.
It is also worth noticing that as the context window size decreases (from 128k to 4k), the necessity of compression increases, and the relative performance gain provided by \method becomes increasingly pronounced.

Moreover, we scrutinize the stability of various existing compression methods under severely tightening token budgets, as visualized in Figure~\ref{fig:Score}.
While strict token constraints typically degrade model outputs, \method exhibits exceptional resilience and consistently maintains a dominant lead over other approaches across all tested token budgets from 2000 down to 500 tokens.
For the results of the \texttt{LLaMA2-7B-chat-4k} model in Figure~\ref{fig:Score}(a), reducing the budget results in minimal performance degradation, with the score decreasing only slightly from 27.56 to 25.13.
Remarkably, while outperforming LLMLingua-2 by 29.88\% under the 2000-token budget, \method's 500-token performance still surpasses the uncompressed baseline of 24.01 and is significantly higher than LLMLingua-2's 21.22 at the much more generous 2000-token budget.
A similar trend is observed in in Figure~\ref{fig:Score}(b) with the \texttt{Longchat1.5-7B-32k} model, the 2000-token results not only surpass the original uncompressed performance of 24.98 but also outperform LLMLingua's 20.69.
In the case of the \texttt{Qwen2.5-7B-Instruct} model in in Figure~\ref{fig:Score}(c), \method outperforms LLMLingua, LongLLMLingua, and LLMLingua-2 by 18.37\%, 21.68\%, and 16.75\% respectively.
These findings underscore the ability of \method to leverage limited resources to preserve essential information, highlighting its practical utility.

\begin{figure*}[t]
	\hspace*{-0.3cm}
	\begin{subfigure}{0.21\textwidth}
		\centering
		\includegraphics[height=3.5cm]{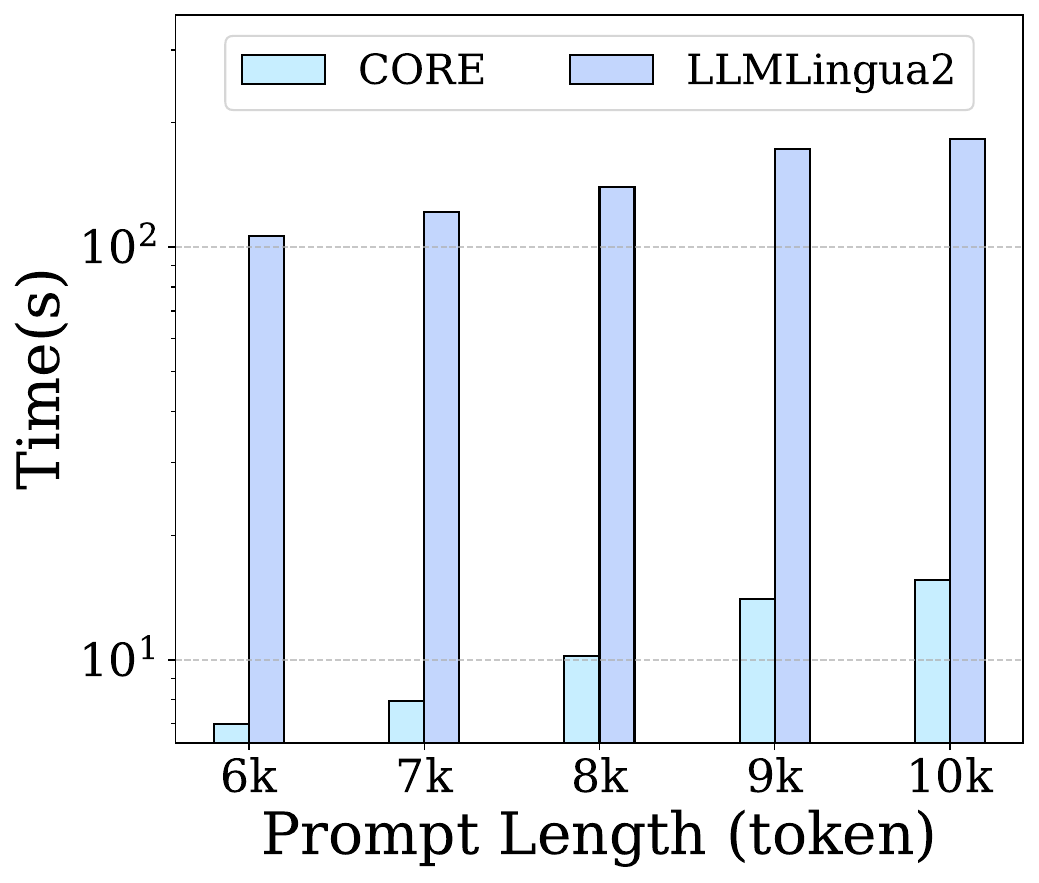}
		\caption{Compression Time}
		\centering
		\label{fig:moblie_time}
	\end{subfigure}
	\hspace{0.03\textwidth}
	\begin{subfigure}{0.21\textwidth}
		\centering
		\includegraphics[height=3.5cm]{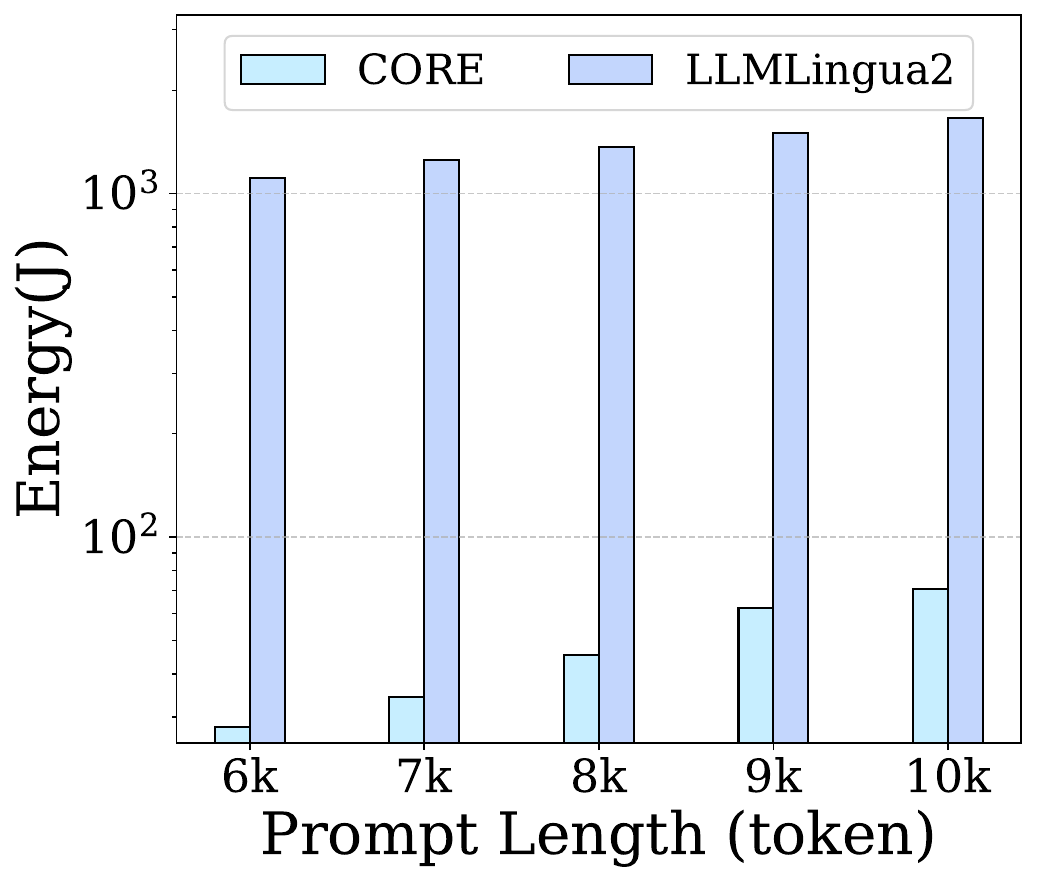}
		\caption{Energy Consumption}
		\label{fig:moblie_energy}
	\end{subfigure}
	\hspace{0.03\textwidth}
	\begin{subfigure}{0.21\textwidth}
		\centering
		\includegraphics[height=3.5cm]{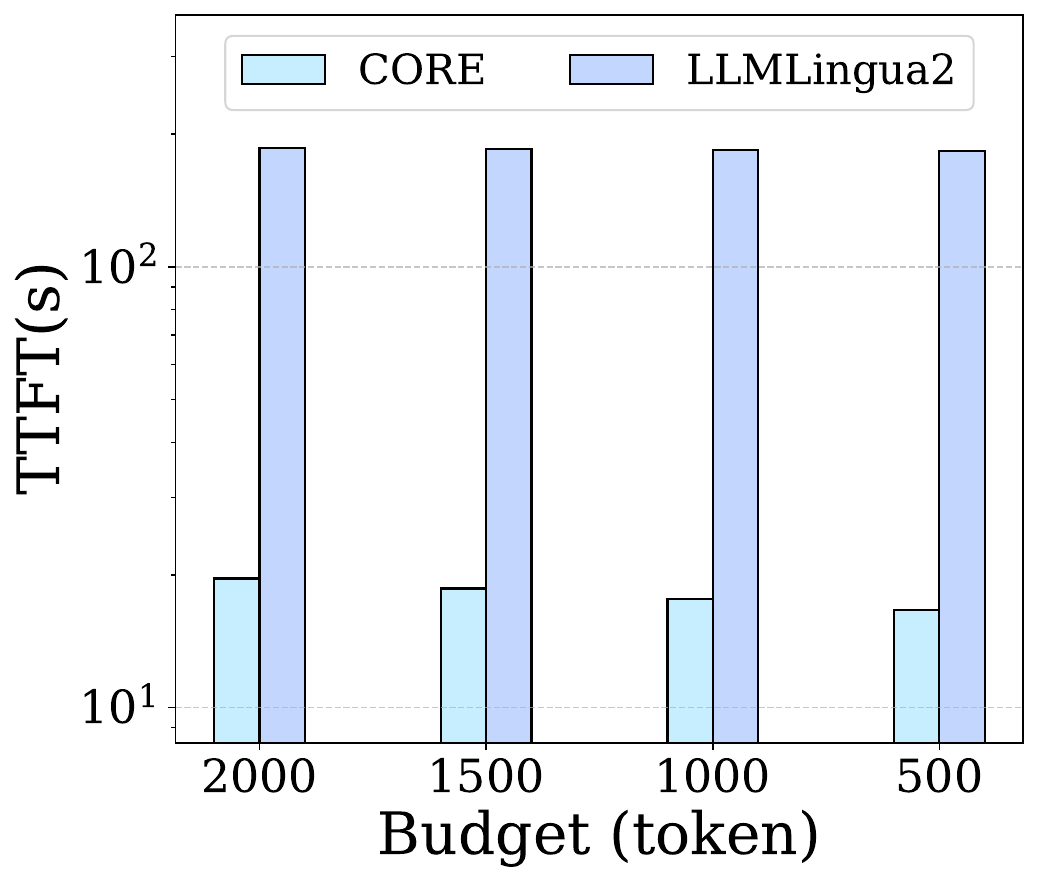}
		\caption{TTFT}
		\label{fig:mobile_ttft}
	\end{subfigure}
	\hspace{0.03\textwidth}
	\begin{subfigure}{0.21\textwidth}
		\centering
		\includegraphics[height=3.5cm]{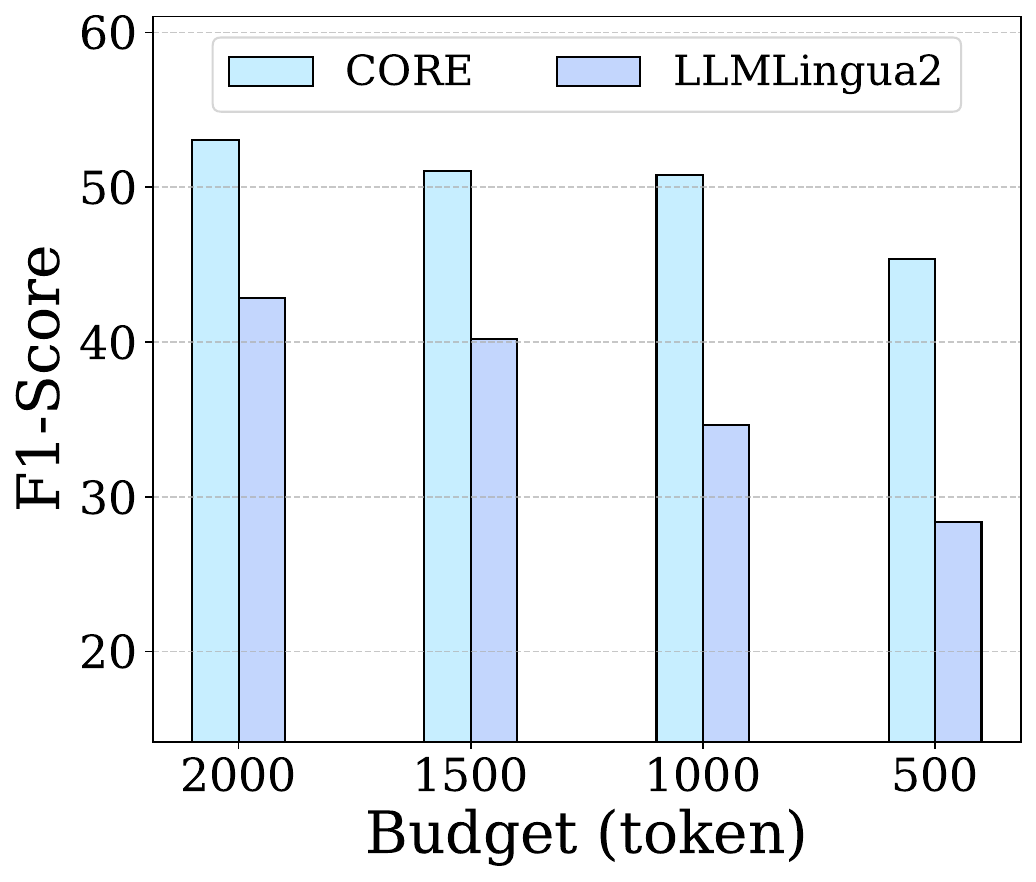}
		\caption{F1-Score}
		\centering
		\label{fig:mobile_f1}
	\end{subfigure}
	\caption{Results of mobile-cloud experiments on the NaturalQuestions dataset.}
	\label{fig:Mobile1}
    \vspace{-0.2cm}
\end{figure*}

\textbf{Memory Overhead.}
Secondly, to assess the feasibility on resource-constrained mobile devices, we evaluate the peak memory consumption of all methods, as illustrated in Figure~\ref{fig:M_L_tau}(b).
\method consistently maintains a minimal memory footprint ranging from merely 0.7 GB to 1.7 GB across all evaluation tasks.
Notably, on the NarrativeQA dataset, \method consumes only 1.38 GB, representing a dramatic reduction of 95.95\% compared to LongLLMLingua and 76.25\% compared to LLMLingua-2.
This significant efficiency demonstrates the practical scalability of \method for processing massive documents directly on edge hardware.

Furthermore, referencing Figure~\ref{fig:M_L_tau}(a), the results elucidate distinct underlying resource consumption patterns driven entirely by the underlying processing mechanisms relative to context length.
For LLMLingua and LongLLMLingua, the memory footprint is dominated by the static overhead of their 7B SLMs.
Because these methods employ a chunking strategy, their memory usage remains consistently high at approximately 34 GB, showing negligible variation even with increasing input length.
Conversely, LLMLingua-2 feeds the entire document context directly into its smaller SLM, so its memory consumption consequently scales proportionally with the input length, rising from 2.58 GB on the shorter Qasper dataset ($\approx$5k tokens) to 5.81 GB on NarrativeQA ($\approx$30k tokens).
In contrast, \method operates at the sentence level, and its memory usage is primarily determined by the number of 
sentences processed rather than the total document length.

\textbf{Latency.}
Finally, we rigorously evaluate the time latency of different methods, examining both the compression time and the TTFT under varying concurrency levels.
As depicted in Figure~\ref{fig:TTFT1}(a), \method exhibits distinct advantages in processing speed, driven by its lightweight components.
Regarding the compression time, baseline methods utilizing heavy SLMs suffer from significant latency.
For instance, processing a 37.5k-token context takes LongLLMLingua 668.34 seconds and LLMLingua 341.24 seconds, creating a prohibitive bottleneck for real-time applications.
In contrast, \method completes the same task in merely 6.77 seconds.
This corresponds to a speedup of 98.99\% over LongLLMLingua, 98.02\% over LLMLingua, and a 48.52\% reduction compared to the efficient LLMLingua-2, validating \method's suitability for latency-sensitive edge environments.

Except for compression time results, we also investigate the concurrency scalability by monitoring TTFT as the number of concurrent requests increases in Figure~\ref{fig:TTFT1}(b).
Benefiting from minimal pre-processing overhead, \method consistently maintains the lowest response latency.
At a concurrency level of 5, \method achieves a TTFT of 2.84 seconds, reducing the latency by approximately 98.32\% compared to LongLLMLingua (183.82s) and 99.13\% compared to LLMLingua (312.36s).
Crucially, against the strongest baseline LLMLingua-2 (5.67s), \method still delivers a 49.91\% improvement.
A critical phenomenon observed in high-concurrency settings, \ie, 20 requests, is the divergence in latency growth rates.
For resource-heavy methods like LLMLingua, the TTFT exhibits near-linear growth  ($\approx$1349s).
Conversely, LLMLingua-2 and \method allow for partial parallel execution of compression tasks on the edge device.
The parallelism ensures a scalable and fluid user experience even with multiple requests.

\subsection{Edge-Cloud Experiment}
In this section, we conduct experiments comparing the two lightweight compression methods, \ie, LLMLingua-2 and \method, on a mobile-cloud deployment scenario.

\textbf{Energy Consumption.}
We first measured the average voltage and current of the two methods during operation, with results summarized in Table~\ref{tab:mobileVA}. 
While the average voltage of \method is slightly lower than that of LLMLingua-2,  \method draws only 917 mA on average, just 39.19\% of the baseline's current. 
This substantial reduction in current, primarily attributed to \method's optimized memory usage, leads to a proportional decrease in active power consumption. 
Therefore, \method reduces overall power consumption by approximately 63.80\% compared to LLMLingua-2, demonstrating a clear advantage in intrinsic energy efficiency.

Consequently, we assess overall energy consumption by jointly considering compression time and power usage. 
As shown in Figure~\ref{fig:Mobile1}(a), \method exhibits substantially lower compression latency. 
When the input prompt length increases from 6k to 10k tokens, \method's compression time rises modestly from 6.99s to 15.63s. 
In contrast, LLMLingua-2's latency scales sharply from 106.18s to 182.27s over the same range, enabling \method to achieve a speedup exceeding 91.42\% at 10k tokens. 
Correspondingly, Figure~\ref{fig:Mobile1}(b) illustrates that \method's energy consumption ranges from 28.00J to 70.58J, markedly lower than that of LLMLingua-2. 
At 10k tokens, \method delivers a substantial energy reduction of 95.74\% relative to LLMLingua-2 (1655.56J). 
These findings demonstrate the superior energy efficiency of \method, underscoring its suitability for deployment on power-constrained mobile devices.

\textbf{TTFT.} Response speed is crucial for interactive mobile applications and services. 
We secondly analyze the end‑to‑end responsiveness of the mobile‑cloud system by measuring the TTFT across various strict token budgets, as shown in Figure~\ref{fig:Mobile1}(c). 
In this setup, the LLM is deployed in the cloud, and its inference process does not compete for computational resources with on‑device compression. 
TTFT is thus primarily determined by the latency of the compression stage on the mobile side. 
Across all budgets, \method consistently achieves lower TTFT. 
For instance, under a restrictive 500‑token budget, \method attains a TTFT of 16.63s, whereas LLMLingua‑2 lags at 183.27s. 
This proportional advantage persists at a higher 2000‑token budget, where \method reaches 19.63s, maintaining a strong 89.47\% improvement over the baseline (186.27s). 
Such significant TTFT reductions highlight \method's capability to enable real‑time mobile‑cloud response, substantially enhancing the overall user experience.

\textbf{F1-Score.}
Finally, we evaluate effectiveness by analyzing F1-Score performance under tightening token budgets, with results shown in Figure~\ref{fig:Mobile1}(d). 
\method consistently achieves markedly higher F1-scores across all evaluated budgets compared to LLMLingua-2. 
At the 2000-token budget, \method attains an F1-score of 50.78, surpassing the baseline value of 42.44 by 19.65\%. 
The performance gap further accentuates as resources become limited. At the most constrained 500-token budget, \method maintains an F1-score of 39.80, yielding a substantial 60.07\% improvement over the baseline score of 24.87. 
This demonstrates \method's superior ability to preserve critical information and predictive accuracy even under severe mobile-side constraints, validating its fundamental strength in content preservation.

\begin{table}[t]
	\caption{Mobile device's average power consumption.}
	\centering
	\resizebox{.8\columnwidth}{!}{%
		\begin{tabular}{c|ccc}
			\toprule
			\textbf{}  & \textbf{AVG Voltage} & \textbf{AVG Current} & \textbf{AVG Power} \\
			\midrule
			\textbf{LLMLingua-2}   & 4389 mV      & 2340 mA   & 10.27 W \\
			\midrule
			\textbf{\method} &   4055 mV       & 917 mA   & 3.72 W \\
			\bottomrule
		\end{tabular}%
	}
    \vspace{-0.1cm}
	\label{tab:mobileVA}
\end{table}

	\section{Ablation Study}
In this section, we conduct comprehensive ablation experiments to validate the critical contribution of each key component in \method.
We begin by examining the effectiveness of individual sequential stages in the evidence refining stage, \ie, the clue refining mechanism and the answer pruning phase, before evaluating the progressive contribution of the entire compression pipeline.

\subsection{Effectiveness of Dynamic Clue Refining}
In the clue refining stage, we adopt a dynamic clue subset selection strategy to replace the Top-$K$ method. 
To evaluate the effectiveness of our proposed method, we conduct experiments on all six LongBench datasets using the \texttt{Qwen2.5-7B-Instruct} model. 
Similar to the preliminary experimental setup in Section~\ref{sec:motivation}, we compare the clue subset obtained via the orthogonal residual retrieval strategy and the elbow method with those of a fixed Top-$K$ semantic similarity approach. 
These clue subsets, together with the candidate answer sets, are fed as context to the LLM.

\begin{figure}[t]
	\centering
	\includegraphics[width=0.95\columnwidth]{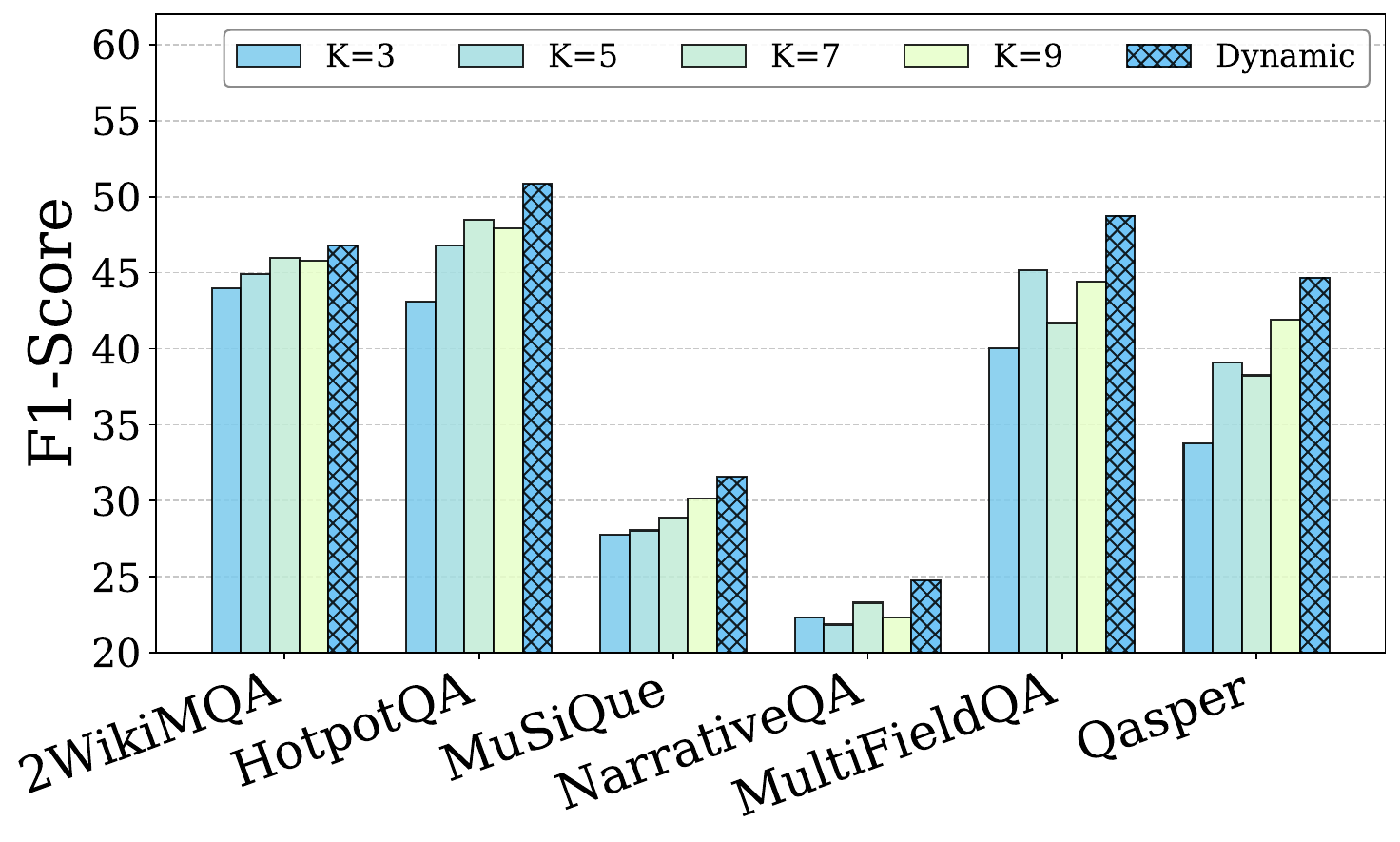}
	\caption{Comparison of F1-scores between dynamic clue subset selection and fixed Top-$K$ approaches across LongBench datasets.}
	\label{fig:clue_ablation}
    \vspace{-0.2cm}
\end{figure}

As illustrated in Figure~\ref{fig:clue_ablation}, the experimental results show that our dynamic method consistently achieves higher F1-scores than all fixed-$K$ configurations across all six datasets. 
Specifically, the dynamic strategy outperforms the best fixed-$K$ baseline by margins ranging from 0.8\% on 2WikiMQA to 3.6\% on MultiFieldQA, with an average improvement of 2.1\% across all datasets. 
Notably, on HotpotQA, our method achieves an F1-score of 50.84, surpassing the best fixed configuration (K=7) by 2.3\%. 
Similarly, on Qasper, the dynamic approach attains 44.67 F1-score, outperforming the best fixed-K baseline (K=9) by 2.8\%.
The advantage becomes more pronounced when $K$ is chosen arbitrarily. 
For instance, on MultiFieldQA, selecting a fixed $K=3$ results in a significant 8.8\% performance drop compared to our dynamic method, while on Qasper, fixed $K=3$ leads to a 10.9\% degradation. 
Conversely, in HotpotQA, where questions typically require synthesizing information from multiple documents, choosing $K=3$ yields a 7.7\% lower F1-score than our adaptive approach. 
In contrast, our method eliminates such guesswork by automatically determining the optimal subset size for each query through the elbow method.

These results demonstrate that by automatically identifying the point of maximum curvature in the relevance score sequence, our method adaptively determines the optimal subset size for each individual query, thereby ensuring comprehensive semantic coverage without introducing excessive noise.

\subsection{Preservation of Ground-Truth Answers}

Building upon the validated clue selection mechanism, we further investigate whether the answer pruning phase can effectively retain ground-truth answers under varying compression ratios.
We formally define the answer preservation rate as the ratio of critical sentences containing ground-truth answers retained after compression to those before compression.

\begin{figure}[t]
\centering
\includegraphics[width=0.95\columnwidth]{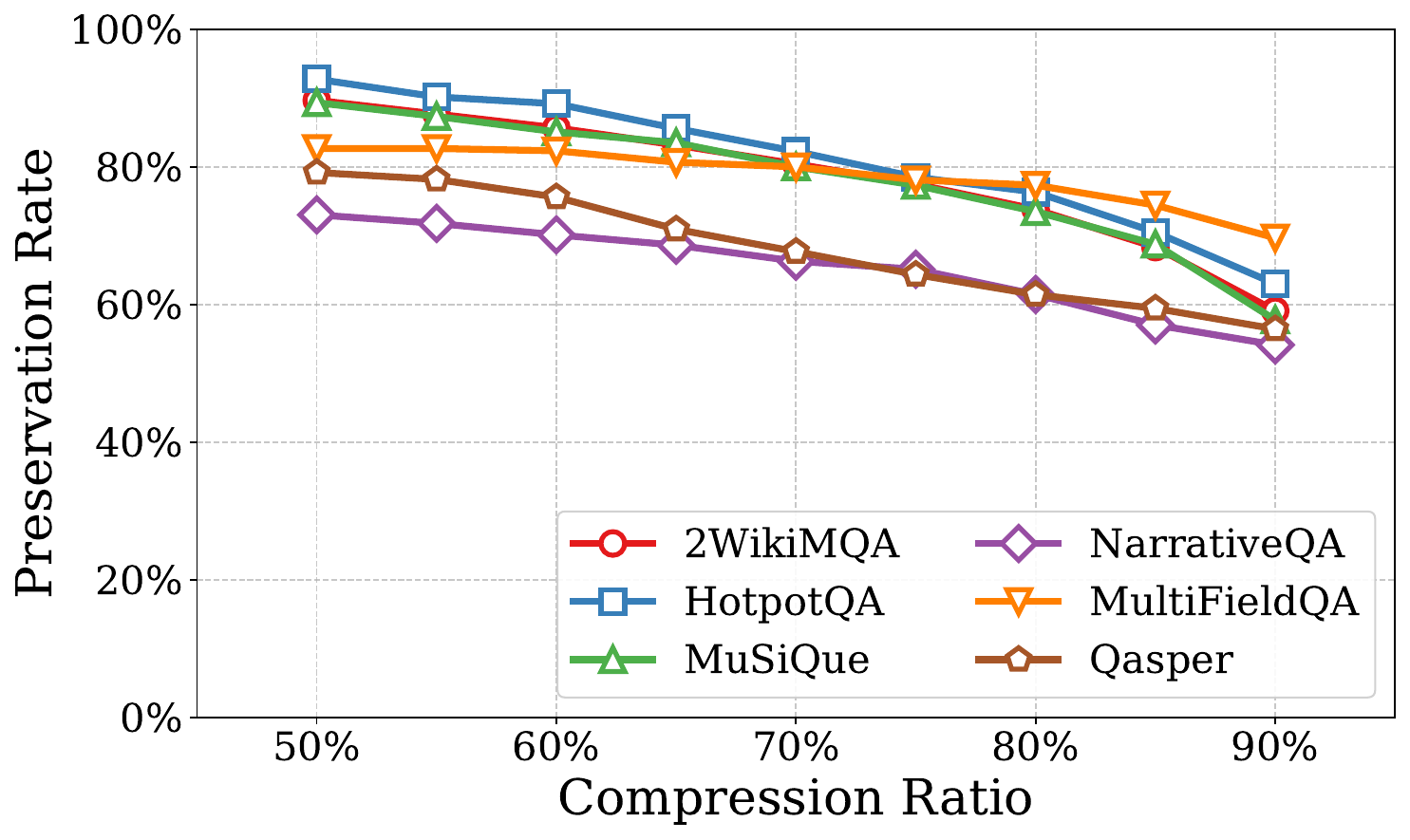}
\caption{Answer preservation rates across six LongBench datasets under different compression ratios.}
\label{fig:answer_preservation}
\vspace{-0.2cm}
\end{figure}

As shown in Figure~\ref{fig:answer_preservation}, we measure the answer preservation rate under compression ratios ranging from 50\% to 90\%. 
It is worth noting that perfect preservation of all ground-truth answer sentences is not strictly necessary for a large language model to generate correct answers. The results indicate that \method achieves a remarkably high preservation rate. 
Under a moderate compression ratio of 50\%, the preservation rates on 2WikiMQA, HotpotQA, and MuSiQue are all around 90\%. 
Even under an aggressive compression ratio of 90\%, \method maintains an average preservation rate of over 60\% for ground-truth answers across all datasets. 
NarrativeQA poses the greatest challenge due to its exceptionally long contexts, which often exceed 30k tokens, yet it still maintains a preservation rate above 54\% under extreme compression. 
Notably, because long contexts frequently contain multiple sentences conveying identical or complementary answer information, retaining every sentence that contains a ground-truth answer is not required for generating correct responses. 
Therefore, this result remains meaningful.

These results demonstrate that the proposed answer importance metric in \method effectively identifies and prioritizes critical answer sentences, thereby robustly ensuring that essential evidence is preserved even when the majority of the context is pruned.

\subsection{Progressive Impact of the Full Pipeline}
\begin{figure}[t]
	\centering
	\includegraphics[height=5.5cm]{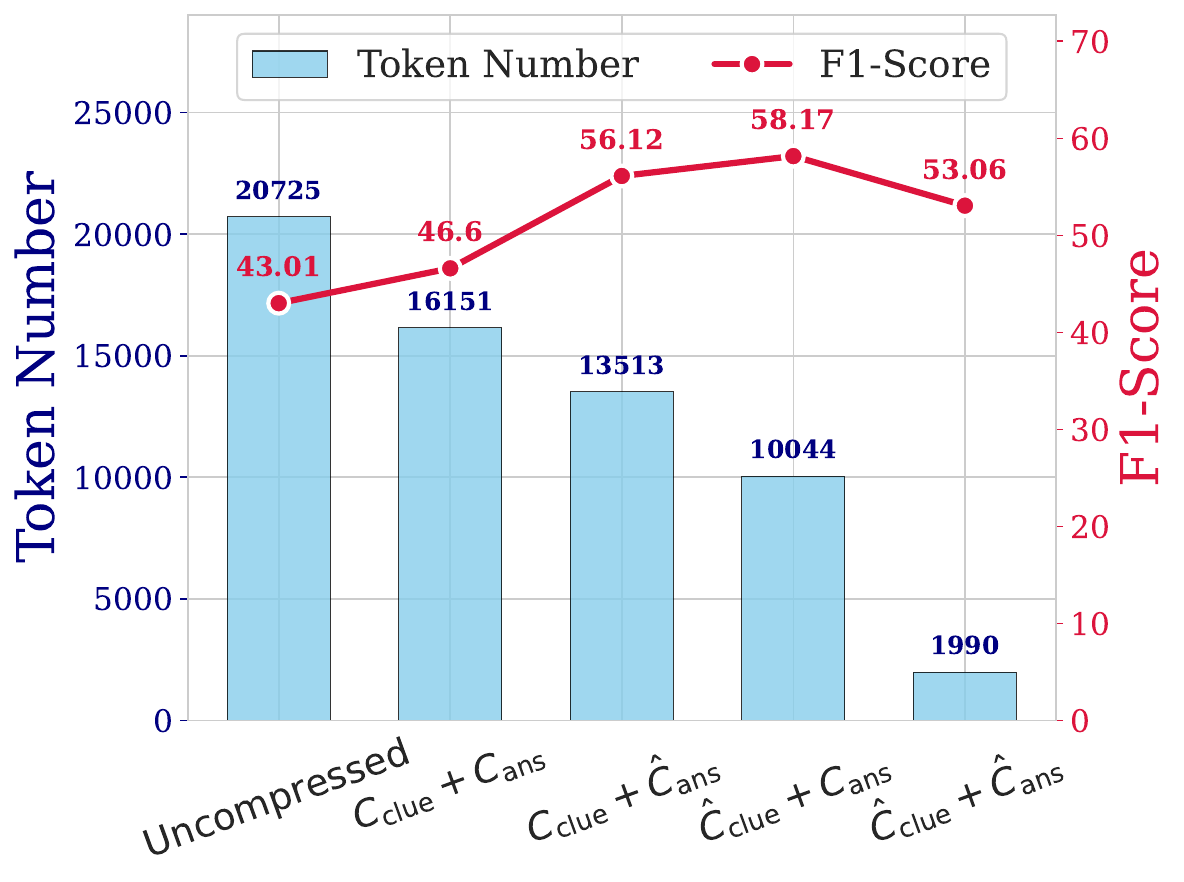}
	\caption{Ablation study results within a 2000-token budget on the NaturalQuestions dataset.}
	\centering
	\label{fig:ablation}
    \vspace{-0.2cm}
\end{figure}

Having established the effectiveness of the individual refining phases, we finally conduct ablation experiments on the NaturalQuestions dataset using \texttt{gpt-3.5-turbo-16k} under a strict 2,000‑token budget to validate the contribution of each stage in \method. 
As illustrated in Figure~\ref{fig:ablation}, we compare five configurations that represent successive stages of the compression process:

\begin{itemize}
    \item \textbf{Uncompressed:} The original context containing 20725 tokens with an F1-score of 43.01, serving as the baseline.
    \item \textbf{$\mathbb{C}_{\text{clue}}+\mathbb{C}_{\text{ans}}$:} The initial filtering stage that constructs candidate clue and answer sets based on semantic similarity and entity recognition. This reduces the input to 16151 tokens while raising the F1-score to 46.60, thereby mitigating information loss caused by exceeding the model's context limit.
    \item \textbf{$\mathbb{C}_{\text{clue}}+\hat{\mathbb{C}}_{\text{ans}}$:} This intermediate configuration applies answer pruning to the candidate sets while retaining all candidate clues without orthogonal refinement. The token count reduces to 13,513 with the F1-score improving to 56.12, demonstrating that filtering irrelevant answers based on spatial proximity to clues alone can effectively enhance reasoning accuracy.
    \item \textbf{$\hat{\mathbb{C}}_{\text{clue}}+\mathbb{C}_{\text{ans}}$:} Building upon the filtered candidates, we apply iterative orthogonal retrieval to distill a salient clue subset while retaining all candidate answers. This configuration achieves 10,044 tokens with an F1-score of 58.17, corresponding to a 24.85\% gain over the results after the first stage.
    \item \textbf{$\hat{\mathbb{C}}_{\text{clue}}+\hat{\mathbb{C}}_{\text{ans}}$:} The jointly optimized configuration that applies both clue refining and answer pruning. This attains the F1-score of 53.06 with just 1990 tokens, confirming that even when compressed far below the model's window capacity, the context maintains sufficient information to support reasoning.
\end{itemize}

Collectively, these results validate that \method's two-stage design provides a principled approach to prompt compression, preserving essential information while achieving aggressive compression ratios suitable for resource-constrained edge deployment.
	
	\section{Conclusion}\label{sec:conclusion}
	In this paper, we propose \method, a lightweight two-stage prompt compression method designed for the QA agent on resource-constrained edge devices.
\method first aggregates candidate clues and answers via semantic similarity and entity analysis, and subsequently performs fine-grained compression to optimize the prompt content within a strict token budget.
Because \method eliminates the reliance on computationally intensive auxiliary SLMs and utilizes efficient lightweight components, it achieves significant reductions in memory overhead and latency while maintaining high generation accuracy.
Experiment results demonstrate that \method improves accuracy by at least 30.19\% compared to the baseline methods, while reducing memory usage and compression latency by at least 50.47\% and 48.52\%, respectively, revealing it is well-suited for real-time QA applications with RAG on edge platforms.

    \section{Related Work}\label{sec:relatedwork}
	\subsection{Long Context Problem}
Long context modeling constitutes a persistent challenge for LLMs based on the Transformer architecture \cite{raiaan2024review, liu2025comprehensive}.
While substantial research has focused on intrinsic model enhancements—ranging from efficient attention mechanisms \cite{shah2024flashattention, zhang2025spargeattn} to context window extension via continued pre-training \cite{zhang2024extending, pawar2024and} and advanced positional encoding \cite{zhang2024found, irani2025positional}.
These methods typically necessitate structural modifications or computationally expensive retraining.
Consequently, they are often inapplicable to scenarios involving black-box LLMs where the model weights are frozen.
Furthermore, recent studies have revealed that even with extended context windows, LLMs often suffer from the ``lost in the middle'' phenomenon \cite{liu2024lost}, where information located in the middle of long contexts tends to be overlooked, fundamentally limiting the effective utilization of increased context lengths.

\subsection{Training-Based Compression}
To address the long-context bottleneck without altering the target LLM, prompt compression has emerged as a viable strategy \cite{li2024prompt}.
Training-based approaches generally involve fine-tuning auxiliary discriminators to assess information importance.
For instance, LLMLingua-2 \cite{pan2024llmlingua} employs a data-driven strategy, distilling knowledge from GPT-4 into a binary classifier based on an SLM to perform token-level retention decisions.
Similarly, CPC \cite{liskavets2025prompt} utilizes contrastive learning to train a sentence-level discriminator, evaluating relative importance through pairwise comparisons.
Beyond discriminative approaches, AutoCompressor \cite{chevalier2023autocompressor} trains models to compress long contexts into compact ``summary tokens'' through soft prompts, while GIST \cite{mu2023gist} distills task-specific instructions into learned gist tokens achieving up to $26\times$ compression.
However, these methods rely on training process and high-quality data, which makes them unsuitable for resourse-constrained edge devices.

\subsection{Training-Free Compression}
Training-free methodologies operate without additional fine-tuning, leveraging intrinsic metrics derived directly from pre-trained models to guide pruning.
Early works like Selective Context \cite{li2023unlocking} utilized self-information as a core metric, employing SLMs to quantify and filter content based on information-theoretical redundancy.
Subsequently, perplexity-based methods gained prominence.
LLMLingua \cite{jiang2023llmlingua} and its long-context variant LongLLMLingua \cite{jiang2024longllmlingua} adopt a coarse-to-fine strategy, iteratively removing tokens that contribute minimal perplexity to the generation process.
More recently, attention-driven mechanisms have attracted interest.
AttnComp \cite{zhao2025leveraging} analyzes self-attention matrices to construct semantic units for pruning, while EHPC \cite{fei2025efficient} identifies and leverages specialized attention heads within SLMs to pinpoint critical tokens.
Despite the diversity of these metrics, current training-free methods face severe deployment hurdles, \ie, heavy memory overhead and high computation latency on resource-constrained edge devices, primarily due to their heavy reliance on auxiliary SLMs.
These constraints underscore the urgent need for a memory-efficient, low-latency compression framework tailored for the edge.

\subsection{Named Entity Recognition}
Named Entity Recognition (NER) is a fundamental technique for extracting key information from unstructured text by classifying mentions into predefined categories such as person names (\(\mathbf{PER}\)), organizations (\(\mathbf{ORG}\)), and locations (\(\mathbf{LOC}\), \(\mathbf{GPE}\)) \cite{keraghel2024survey, nasar2021named}.
While pre-trained language models like BERT \cite{devlin2019bert} have enabled context-aware entity recognition, their large size often exceeds edge device constraints.
Consequently, lightweight solutions such as spaCy \cite{honnibal2020spacy} and DistilBERT \cite{sanh2019distilbert} have gained traction for resource-constrained deployment, offering compact architectures (\eg, occupying $\approx$ 20 MB for model weights) while retaining substantial accuracy.
Moreover, NER serves as a critical component in retrieval-augmented generation systems for identifying key entities that bridge queries with relevant context \cite{jimenez2024hipporag}, motivating our approach of utilizing lightweight entity recognition to construct semantically meaningful candidate sets for compression.

	\bibliographystyle{IEEEtran}
	\bibliography{contents/refs}

	\begin{IEEEbiography}[{\includegraphics[width=1in,height=1.2in,clip,keepaspectratio]{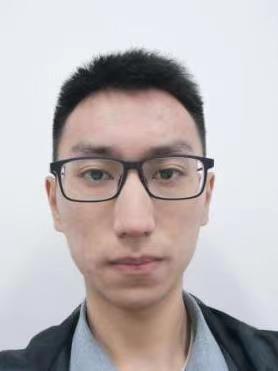}}]
{Zihuai Xu} received the B.S. degree from the School of Computer Science and Technology at Northwestern Polytechnical University (NWPU) in 2021 and the M.S. degree from the same school in 2024. He is currently pursuing the Ph.D. degree in the School of Computer Science and Technology at the University of Science and Technology of China (USTC). His research interests primarily focus on edge computing and large language models.
\end{IEEEbiography}

\begin{IEEEbiography}[{\includegraphics[width=1in,height=1.2in,clip,keepaspectratio]{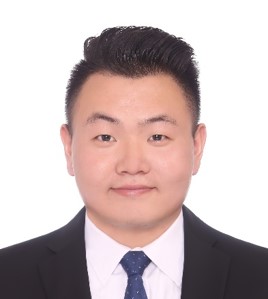}}]{Yang Xu}(Member, IEEE) is currently an associate researcher in the School of Computer Science and Technology at University of Science and Technology of China. He got his Ph.D. degree in computer science and technology from University of Science and Technology of China in 2019. He got his B.S. degree in Wuhan University of Technology in 2014. His research interests include Ubiquitous Computing, Deep Learning and Mobile Edge Computing.
\end{IEEEbiography}

 \begin{IEEEbiography}
[{\includegraphics[width=1in,height=1.2in,clip,keepaspectratio]{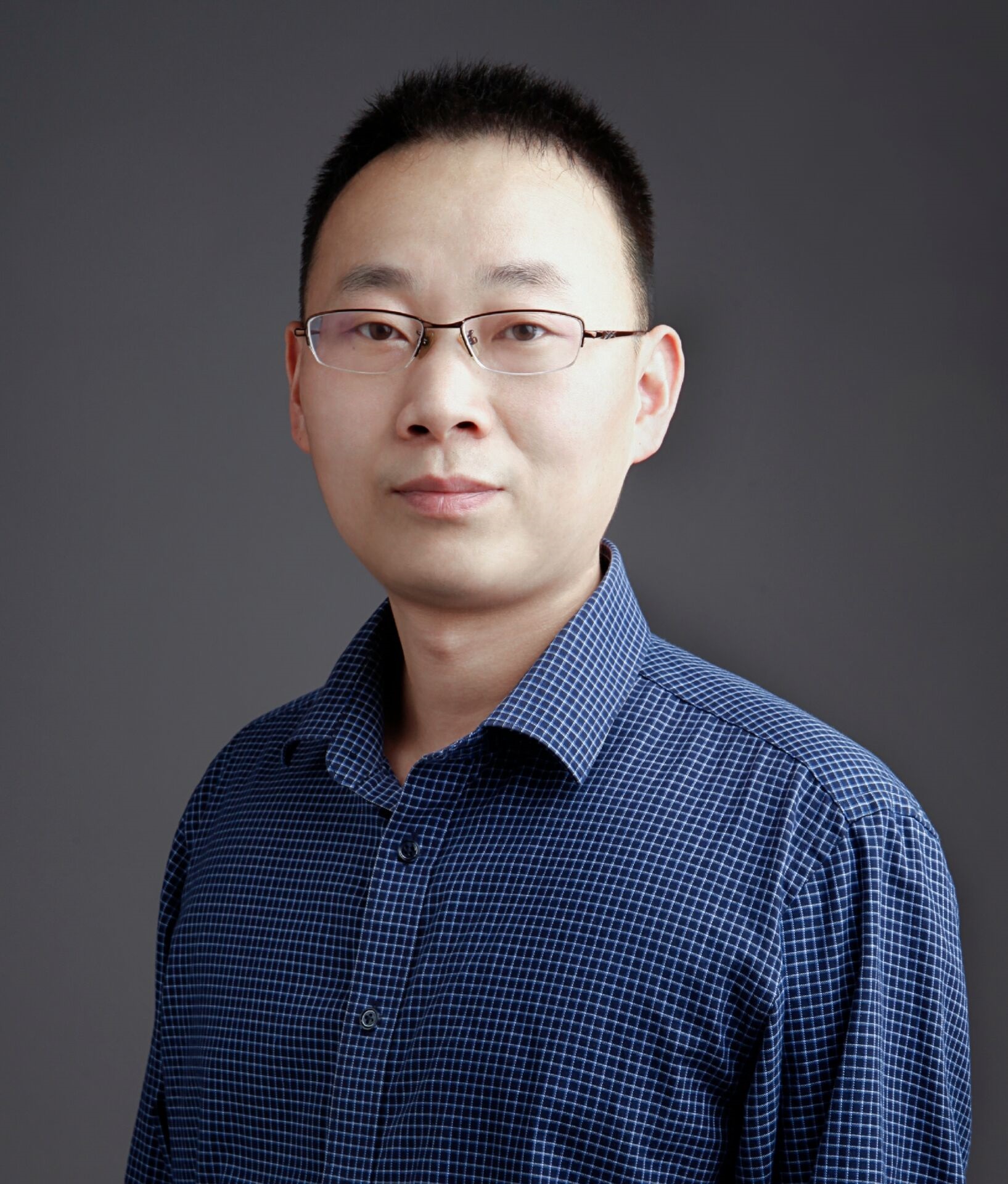}}]
 {Hongli Xu}
 (Member, IEEE) received the B.S. degree in computer science from the University of Science and Technology of China in 2002, and the Ph.D. degree in computer software and theory from the University of Science and Technology of China in 2007. 
 He is a professor at the School of Computer Science and Technology, University of Science and Technology of China, China. He was awarded the Outstanding Youth Science Foundation of NSFC, in 2018. He has won the best paper award or the best paper candidate at several famous conferences. 
 He has published more than 100 papers in famous journals and conferences, including the IEEE/ACM Trans. on Networking, IEEE Trans. on Mobile Computing, Infocom and ICDE, etc. 
 His main research interests are software software-defined networks, edge computing, and Internet of Things.
 \end{IEEEbiography}
 
\begin{IEEEbiography}[{\includegraphics[width=1in,height=1.2in,clip,keepaspectratio]{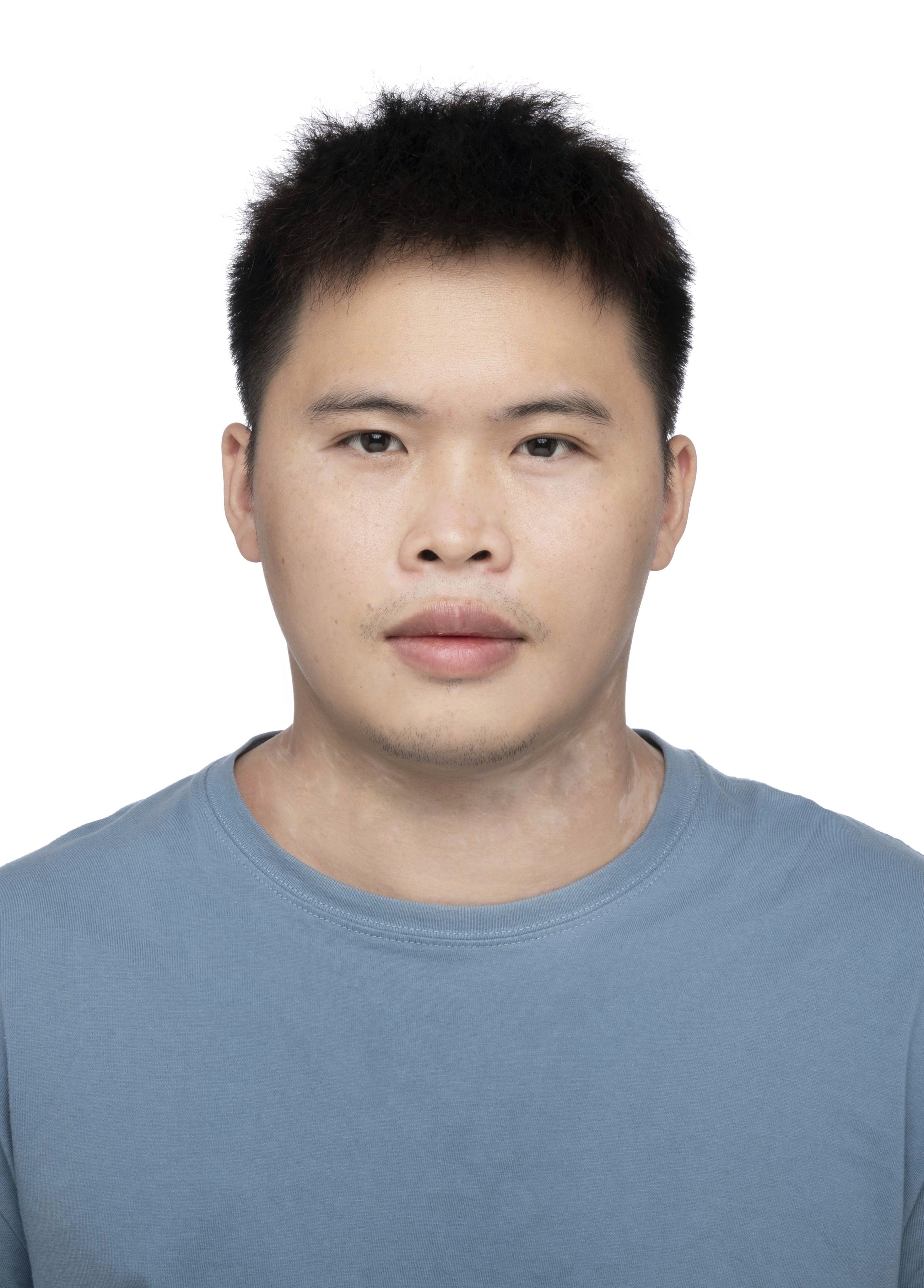}}]
{Yunming Liao} received the B.S. degree in 2020 from the University of Science and Technology of China. He is currently pursuing his Ph.D. degree in the School of Computer Science and Technology, University of Science and Technology of China. His research interests include mobile edge computing and federated learning. 
\end{IEEEbiography}

\begin{IEEEbiography}
  [{\includegraphics[width=1in,height=1.25in,clip,keepaspectratio]{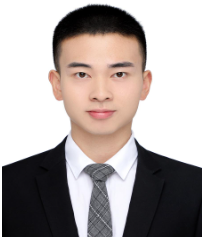}}]
	{Zhiwei Yao}
	received the B.S. degree in 2021 from the Hohai University. He is currently pursuing his Ph.D. degree in the School of Computer Science and Technology, University of Science and Technology of China. His main research interests are edge computing and federated learning. 
\end{IEEEbiography}

\begin{IEEEbiography}[{\includegraphics[width=1in,height=1.2in,clip,keepaspectratio]{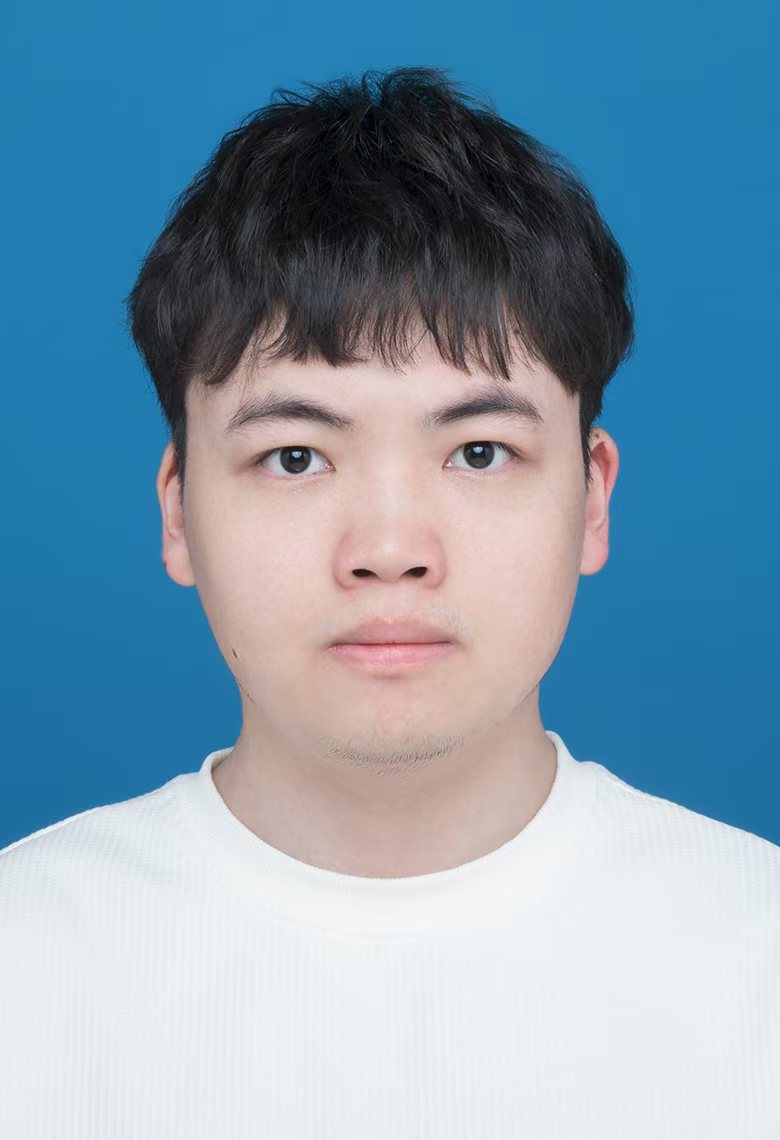}}]
{Zuan Xie} received the B.S. degree from the School of Computer Science and Technology from University of Science and Technology of China (USTC) in 2022. He is currently working toward the Ph.D. degree in the School of Computer Science and Technology, USTC. His main research interests include edge computing, federated learning, and distributed machine learning.
\end{IEEEbiography}

\end{document}